%% file: acl_latex.tex
\documentclass[11pt]{article}

\usepackage[final]{acl}

\usepackage{times}
\usepackage{latexsym}

\usepackage[T1]{fontenc}

\usepackage[utf8]{inputenc}

\usepackage{microtype}

\usepackage{inconsolata}

\usepackage{graphicx}

\title{OmniDecisionBench: Evaluating and Benchmarking the Multimodal Jev-Style System One Decision Models}

\author{Tianxiang Gao$^{1}$ \quad
        Jinzhe Li$^{1,3}$ \quad
        Yue Zhou$^{1}$ \quad
        Dandan Guo$^{1}$ \\
        \textbf{Qi Wang$^{1}$ \quad
        Diana Inkpen$^{4}$ \quad
        Yuan Wu$^{1}$\thanks{Corresponding author} \quad
        Yi Chang$^{1,2,3}$} \\
        $^{1}$School of Artificial Intelligence, Jilin University \\
        $^{2}$Engineering Research Center of Knowledge-Driven Human-Machine Intelligence, Jilin University \\
        $^{3}$International Center of Future Science, Jilin University \\
        $^{4}$University of Ottawa\\
        \{gaotx5524, jinzhe25, yuezhou24\}@mails.jlu.edu.cn, \{qiwang, guodandan, yuanwu, yichang\}@jlu.edu.cn
}

\usepackage{amsmath,amssymb,booktabs,array,colortbl}
\definecolor{mmBest}{RGB}{218,236,232}
\definecolor{mmBand}{RGB}{238,243,246}
\newcommand{\bestcell}[1]{\cellcolor{mmBest}\textbf{#1}}
\newcommand{\secondcell}[1]{\underline{#1}}
\newcommand{\sectrow}[2]{\multicolumn{#1}{>{\columncolor{mmBand}}l}{\textit{#2}}}

\begin{document}
\maketitle
\begin{abstract}
\input{section/abstract}
\end{abstract}
\input{section/introduction}
\input{section/related_work}
\input{section/preliminaries}
\input{section/benchmark}
\input{section/experiments}
\input{section/conclusion}
\input{section/limitations}
\input{section/ethics}
\bibliography{references}
\clearpage
\appendix
\input{section/appendix}
\end{document}

%% file: section/abstract.tex
Multimodal System One models answer visual tasks with typed decisions and native probabilities, yet no holistic evaluation compares them with general VLMs or maps either class's capability boundaries. We introduce \textbf{OmniDecisionBench}, a holistic benchmark of multimodal System One models. Its main evaluation assigns 45 datasets (192,975 records) to five capability families and twelve primary angles across four typed interfaces, scoring accuracy and candidate-probability calibration. A validation layer adds three single-factor diagnostics varying visual presentation, textual evidence sufficiency, and target dataset. Across three System One models and two general VLMs, three regularities emerge. First, strengths depend on angle and metric: System One models lead discrete tasks overall with lower $\mathrm{ECE}_{15}$ on most perception, reading, and reasoning angles (Jev~1.14: 80.14\% accuracy, 4.50\% $\mathrm{ECE}_{15}$), while general VLMs lead ordinal grading and outperform Jev~1.14 among the three models tested on direct Box grounding. Second, multi-image and abstract reasoning bound every model, where confidence rather than accuracy separates them. Third, reliability shifts with the input: rendering only options nearly triples Jev~1.14's selective risk at nearly flat mean signed bias, and donor risk cutoffs often exceed target risk across unseen datasets and task families. \textbf{OmniDecisionBench} thus maps which model class suits each angle and where reliability ends.
The code and dataset will be available at \url{https://github.com/Glax147/OmniDecisionBench}.

%% file: section/introduction.tex
\section{Introduction}
\label{sec:introduction}
Real-world visual workflows routinely require fast, executable decisions rather than open-ended narration: extracting a document field \citep{mathew2021docvqa}, verifying a spatial claim \citep{liu2023vsr}, grading physical damage \citep{alam2020crisis}, or localizing a referred object \citep{yu2016refcoco}. Two model classes now serve these decisions. Structured \emph{System One} decision models answer through typed output interfaces---Choice for finite selection, Noul for binary verification, Score for ordered grading, and Box for bounding coordinates---paired with native predictive probabilities \citep{typesafe2026models}, while general-purpose VLMs reach the same decisions through generic instruction following. Choosing between them requires two answers: which class is more \emph{capable} on a given kind of visual task, and where its \emph{decision reliability} ends, that is, where its probabilities stop supporting deferral, where a changed layout breaks it, and where a tuned risk threshold stops transferring.

For multimodal inputs, the capability boundaries of both model classes remain unmapped. Text-domain evaluations have mapped the capability spectrum, calibration, selective prediction, and decision reliability of System One models \citep{deusser2026jev,li2026soarena}, but never see an image. Holistic VLM suites cover many visual angles, often through multiple-choice questions \citep{lee2024vhelm,liu2023mmbench,fu2023mme,li2023seedbench}, but evaluate general VLMs only and score whether the chosen answer is correct, not the native probability behind it. Visual inputs change both axes: evidence spans typography \citep{liu2024ocrbench}, chart geometry \citep{masry2022chartqa}, cross-view scenes \citep{yang2025mmsi}, abstract diagrams \citep{xu2025visulogic}, and natural photography \citep{hudson2019gqa}, and the same question can change in accuracy, error ranking, and evidence use once its text moves into rendered pixels. What is missing is a holistic evaluation that places both model classes on identical visual records and locates where each one's capability and reliability end.

\begin{figure*}[t]
\centering
\includegraphics[width=\textwidth]{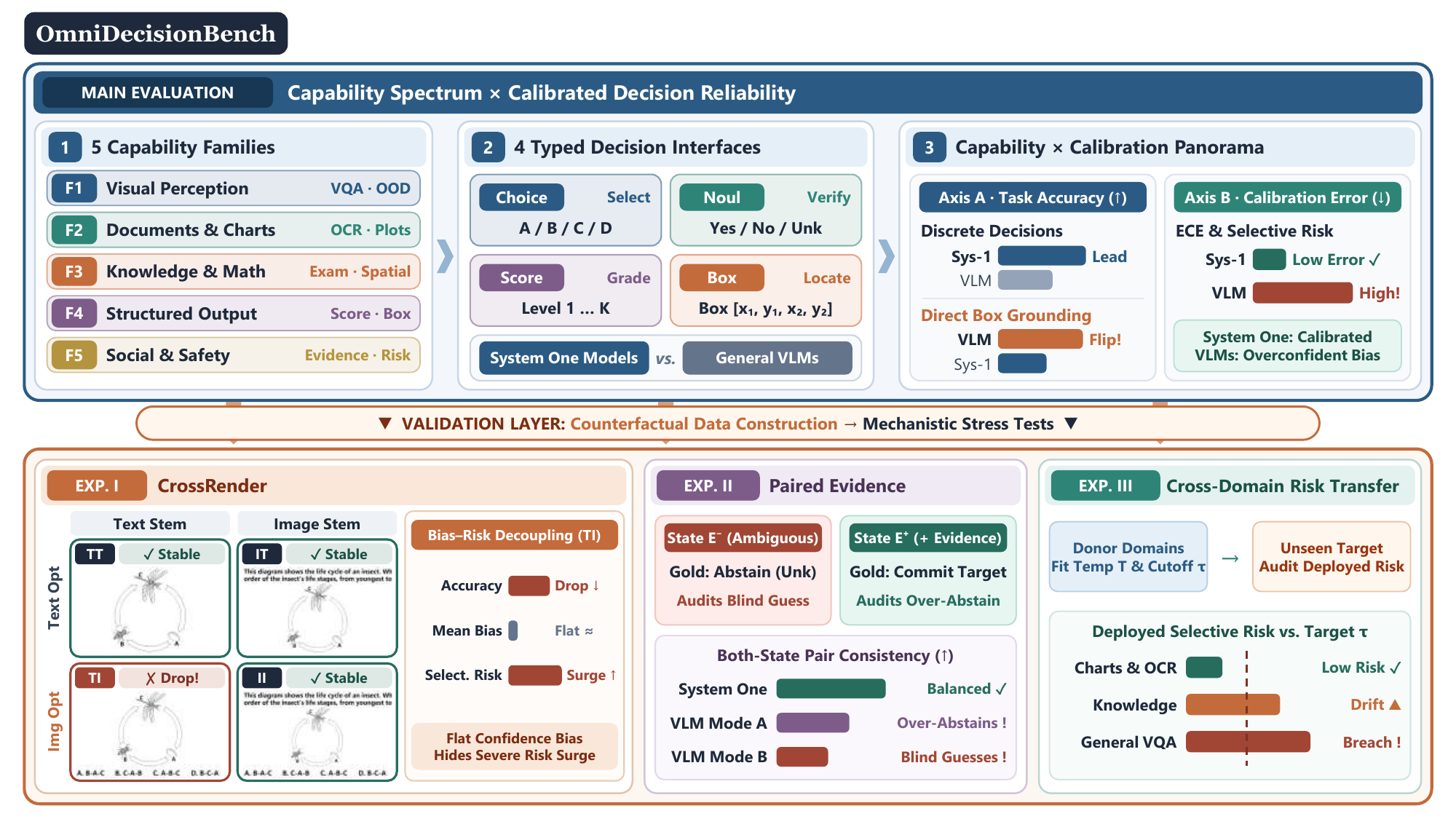}
\caption{Overview of \textbf{OmniDecisionBench} for holistic benchmarking in multimodal System One models. \textbf{Top, Main Evaluation:} five capability families F1--F5 across 45 datasets and 192,975 records are evaluated over four typed decision interfaces---Choice, Noul, Score, and Box---to chart the capability and calibration panorama. \textbf{Bottom, Validation Layer:} three single-factor stress tests probe stem-by-option presentation in Exp.~I \textsc{CrossRender}, paired textual evidence $E^-$ versus $E^+$ in Exp.~II, and cross-dataset risk transfer against a donor cutoff $\tau$ in Exp.~III.}
\label{fig:overview}
\end{figure*}

We introduce \textbf{OmniDecisionBench}, overviewed in Figure~\ref{fig:overview}, as a two-layer benchmark for capability and decision reliability in multimodal System One models. Following HELM and VHELM \citep{liang2023helm,lee2024vhelm}, the \textbf{main evaluation} assigns 45 public datasets totaling 192,975 formal records to five capability families and twelve primary evaluation angles, D1--D12, under frozen manifests so each dataset counts once in macro averages across Choice, Noul, Score, and Box. The \textbf{validation layer} adds three single-factor reliability diagnostics varying stem and option rendering in \textsc{CrossRender}, disambiguating textual context under fixed images, and the target dataset under source-disjoint LODO risk transfer. We compare three System One models---hosted Jev~1.14 \citep{typesafe2026models}, open-weight JEV-27B-VL \citep{autotrust2026deciderprovision27b}, and d3-27B \citep{d3_2026}---with two general VLMs, Qwen3.8-27B \citep{qwen2026qwen3827b} and Gemma~4~31B~IT \citep{google2026gemma431bit,gemmateam2026gemma4}, on identical source images, questions, and labels under each model's documented preprocessor.

We find that System One models and general VLMs have complementary strengths, with capability and reliability ending at clear task boundaries:
\textbf{(i)~Model advantages vary by angle, interface, and metric.} Across discrete tasks, System One models lead overall and achieve lower raw $\mathrm{ECE}_{15}$ on most perception, OCR, chart, and reasoning angles, with Jev~1.14 reaching 80.14\% macro accuracy and 4.50\% $\mathrm{ECE}_{15}$. Conversely, general VLMs lead ordinal grading, social-evidence QA, and direct Box spatial grounding among the three models tested, where Qwen3.8-27B and Gemma~4~31B~IT reach 87.34\% and 82.94\% versus 73.90\% for Jev~1.14.
\textbf{(ii)~Multi-image and abstract reasoning bound every model.} All systems drop on these angles, where general VLMs become severely overconfident while Jev~1.14 curbs high-confidence errors by avoiding confidence $\ge 0.90$ when uncertain.
\textbf{(iii)~Reliability shifts with layout, textual context, and target dataset.} In \textsc{CrossRender}, rendering only options nearly triples Jev~1.14's fixed-coverage risk at nearly flat mean signed confidence bias; aggregate D11/D12 accuracies hide abstention and class-recall trade-offs; and under source-disjoint LODO, donor-tuned 5\% risk cutoffs often exceed target risk across unseen datasets and task families.

Our contributions are threefold:
\begin{itemize}
\setlength{\itemsep}{0.5pt}\setlength{\parskip}{0pt}\setlength{\topsep}{2pt}
\item \textbf{A frozen MECE benchmark:} 45 datasets in a 5-family $\times$ 12-angle primary hierarchy across four typed interfaces, with unified candidate-probability scoring.
\item \textbf{A per-angle capability and calibration map:} a controlled comparison of System One models and general VLMs locating where each class leads and where confidence remains reliable.
\item \textbf{Single-factor reliability diagnostics:} \textsc{CrossRender}, 1,640 MultiBBQ textual-evidence pairs, and 42-dataset source-disjoint LODO risk transfer.
\end{itemize}

%% file: section/related_work.tex
\section{Related Work}
\label{sec:related}
\paragraph{System One decision models.}
System One decision models replace free-form generation with typed decision interfaces and return a native probability for every admissible answer \citep{typesafe2026models}. \citet{deusser2026jev} evaluated text-only Jev across 37 benchmarks on accuracy, calibration, selective prediction, and sequential querying, while \citet{li2026soarena} introduced SO-Arena to chart the capability spectrum and decision reliability, including option-order stability and missing-option discrimination, of text System One models across 5 families and 11 domains. Both studies operate on text inputs. How these models behave when the evidence is visual, when the interface asks for an ordinal grade or a bounding box, and when the same content arrives as text or as rendered pixels has not yet been measured.

\paragraph{Holistic and multimodal evaluation.}
HELM established holistic evaluation of language models across many scenarios and metrics beyond accuracy \citep{liang2023helm}, and VHELM extended this approach to vision--language models across aspects such as perception, reasoning, fairness, multilinguality, and safety \citep{lee2024vhelm}. Targeted VLM suites probe shortcut-resistant reasoning \citep{yue2024mmmupro,chen2024mmstar}, visual induction \citep{xu2025visulogic,tortei2025visres,guo2026visreason}, multi-image geometry \citep{yang2025mmsi,park2026mvh}, OCR \citep{liu2024ocrbench}, referring grounding \citep{yu2016refcoco}, multilinguality \citep{salazar2025kaleidoscope}, and societal evidence or safety \citep{chen2026multibbq,khazaie2026fairlens,wang2025mmsafeaware,degiorgis2026mquest,liu2026mtmcs}. Many of these suites use multiple-choice formats, yet they evaluate general VLMs and score only the correctness of the selected answer; none compares VLMs with System One models or jointly scores typed decision interfaces and the native probabilities that accompany them.

\paragraph{Calibration and selective prediction.}
Temperature scaling is the standard post-hoc correction for miscalibrated networks \citep{guo2017calibration}, and selective prediction turns confidence into an accept-or-defer rule with a target risk \citep{geifman2017selective}. Because calibration and selective risk respond differently to scaling and distribution shift, a calibration score on one dataset and format cannot certify a deployable risk policy. OmniDecisionBench bridges these threads by comparing System One models and general VLMs under one multimodal MECE taxonomy with four typed interfaces and single-factor reliability stress tests in Section~\ref{sec:methods}.

%% file: section/preliminaries.tex
\section{Methods}
\label{sec:methods}
OmniDecisionBench measures two pillars on the same frozen records. The capability spectrum asks how often a model is right on each angle and interface; the reliability layer asks whether its confidence remains usable for deferral when one factor changes at a time. Section~\ref{sec:preliminaries} formalizes the decision instance and both pillars, and Section~\ref{sec:taxonomy} onward describes the taxonomy, the model population, and the diagnostic protocols.

\subsection{Problem Formulation}
\label{sec:preliminaries}
Each decision instance pairs single- or multi-image inputs, along with flagged text-only controls in ScienceQA, with a query, a typed output space, and one of four interfaces: Choice, Noul, Score, or Box; full manifest details appear in Appendix~\ref{sec:execution}. A model returns a typed answer and, on discrete interfaces, candidate probabilities given by native interface probabilities for System One models and normalized conditional likelihoods over valid option codes for general VLMs. The top-label probability serves as confidence, and an answer is correct if it lies in the accepted gold set.

\paragraph{Pillar 1: Capability Spectrum.}
Dataset accuracy is the share of correct answers over all formal records of a dataset. Score uses exact-grade accuracy, and Box counts a prediction as correct at $\mathrm{IoU}\ge 0.5$ in original image coordinates, with malformed or zero-area boxes counted as errors. Angle-level and interface-level capability average datasets equally, and we keep these dataset-macro rates distinct from record-pooled rates.

\paragraph{Pillar 2: Decision Reliability.}
Reliability is assessed along four complementary axes:
(i)~\emph{probabilistic scoring and calibration}, using negative log-likelihood NLL, Brier score, 15-bin top-label expected calibration error $\mathrm{ECE}_{15}$, signed confidence bias defined as mean confidence minus accuracy, and the ranked probability score for ordinal Score;
(ii)~\emph{selective risk}, using tie-aware AURC, the 50\%-coverage error rate $R_{50}$, and the high-confidence error rate $\mathrm{Risk}@90$ at confidence $\ge 0.90$;
(iii)~\emph{presentation robustness} under \textsc{CrossRender}, using the $2\times 2$ accuracy interaction and the shift in signed bias between conditions; and
(iv)~\emph{cross-dataset risk transfer} under source-disjoint LODO policies that target a nominal donor risk. Appendix~\ref{sec:metrics} gives all formulas and floor sensitivity.

%% file: section/benchmark.tex
\subsection{MECE Taxonomy: 5 Families \texorpdfstring{$\times$}{x} 12 Angles}
\label{sec:taxonomy}
OmniDecisionBench comprises 192,975 formal records from 45 public multimodal datasets across 59 dataset--interface slices. Choice dominates by volume, while Noul, Score, and Box cover binary verification, ordinal grading, and direct grounding, with per-interface counts in Table~\ref{tab:overall}. Every item is frozen with its source components, image hashes, gold sets, and transformation version, so all models answer identical records as detailed in Appendix~\ref{sec:execution}.

\begin{figure}[t]
\centering
\includegraphics[width=1.00\columnwidth]{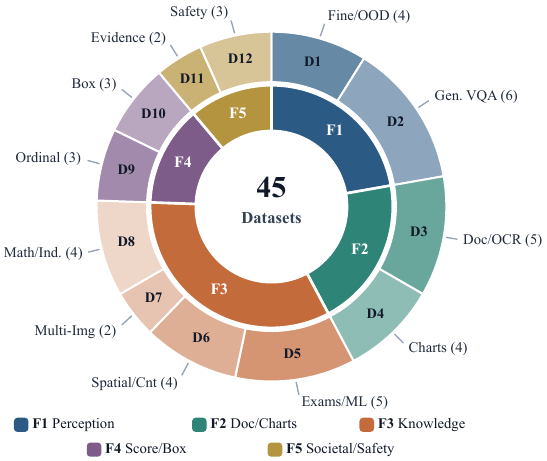}
\caption{MECE hierarchy of OmniDecisionBench: 45 datasets totaling 192,975 records across five families F1--F5 and twelve angles D1--D12; full breakdown in Table~\ref{tab:taxonomy45}.}
\label{fig:taxonomy}
\end{figure}

Following \citet{li2026soarena}, Figure~\ref{fig:taxonomy} and Table~\ref{tab:taxonomy45} assign all 45 datasets to a mutually exclusive, collectively exhaustive MECE primary hierarchy so no dataset is double-counted in macro averages, without implying disjoint underlying skills:
\begin{itemize}
\setlength{\itemsep}{0.5pt}\setlength{\parskip}{0pt}\setlength{\topsep}{2pt}
\item \textbf{F1.~Visual Perception \& Recognition, 10 datasets:} \emph{D1~Fine-Grained \& OOD} covers Flowers102, Food101, and ImageNet-A/R; \emph{D2~General Perception \& VQA} covers VizWiz, MMBench, MME, SEED-Bench, RealWorldQA, and GQA.
\item \textbf{F2.~Document, OCR \& Chart Parsing, 9 datasets:} \emph{D3~Document \& Scene OCR} covers DocVQA, OCRBench, TextVQA, OCR-VQA, and InfographicVQA; \emph{D4~Charts \& Diagrams} covers DVQA, ChartQA, PlotQA, and AI2D.
\item \textbf{F3.~Multimodal Knowledge \& Reasoning, 15 datasets:} \emph{D5~Knowledge, Exams \& Multilingual} covers ScienceQA, A-OKVQA, MMStar, MMMU-Pro, and Kaleidoscope; \emph{D6~Spatial Logic \& Counting} covers CLEVR, NLVR2, VSR, and TallyQA; \emph{D7~Multi-Image \& Cross-View} covers MVH-Bench and MMSI-Bench; \emph{D8~Visual Math \& Abstract Induction} covers MathVista, VisReason, VisRes-Bench, and VisuLogic.
\item \textbf{F4.~Ordinal \& Spatial Interfaces, 6 datasets:} \emph{D9~Ordinal Visual Grading} evaluates WIDER FACE occlusion, Crisis, and BRACOL via Score \citep{yang2016widerface,alam2020crisis,esgario2020bracol}; \emph{D10~Direct Spatial Grounding} evaluates RefCOCO/+/g via Box \citep{yu2016refcoco}.
\item \textbf{F5.~Societal Evidence, Fairness \& Safety, 5 datasets:} \emph{D11~Social Evidence \& Fairness} covers MultiBBQ and FairLens; \emph{D12~Toxicity \& Safety} covers M-QUEST, MMSafeAware, and MTMCS-Bench.
\end{itemize}
Datasets with both Choice and Noul slices are pooled within the dataset before equal-dataset averaging, so a tiny slice cannot distort its angle.

\subsection{Frozen Model Population and Interfaces}
\begin{figure*}[t]
\centering
\includegraphics[width=0.95\textwidth]{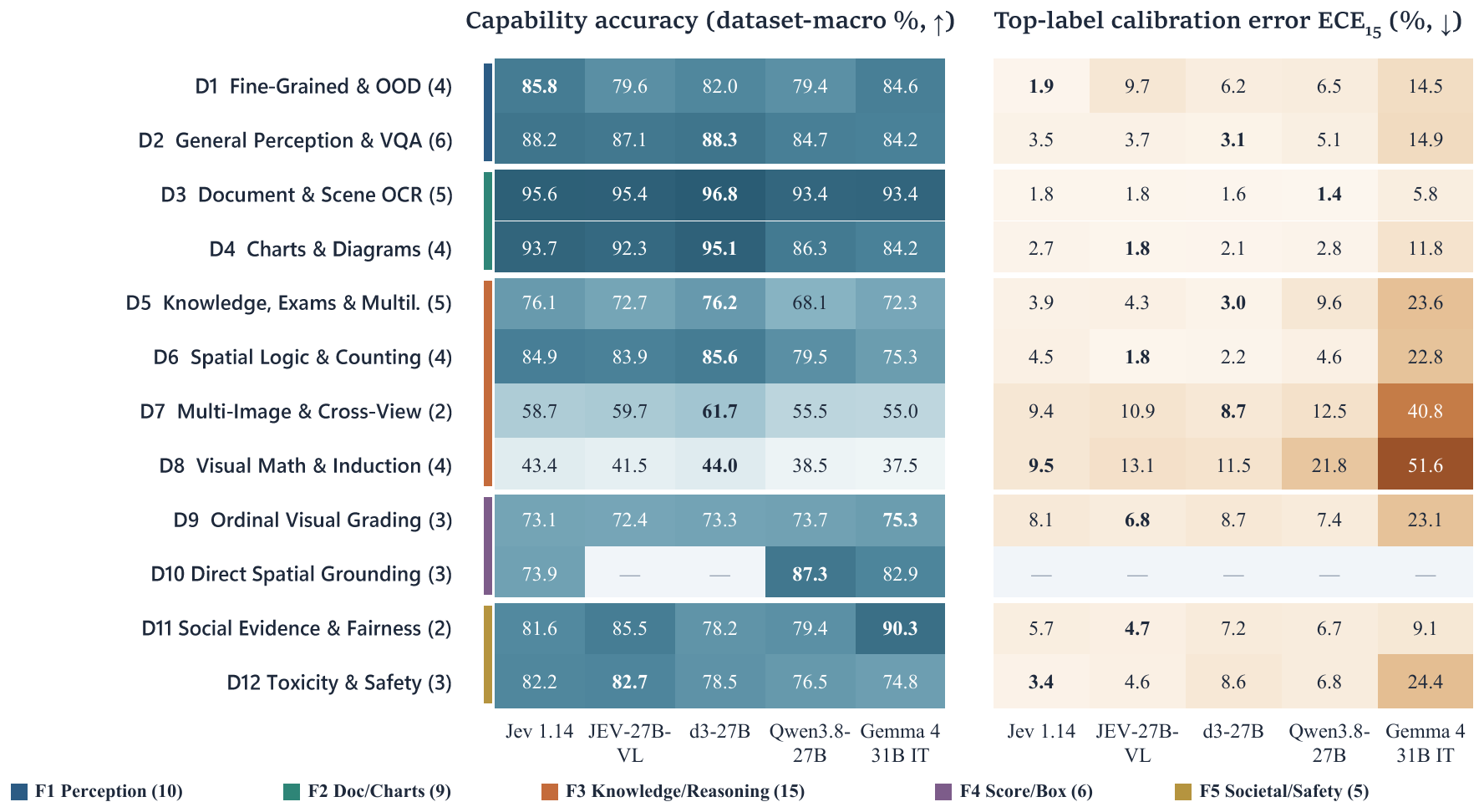}
\caption{Capability and calibration panorama across F1--F5 and D1--D12 over 45 datasets and 192,975 records. Left: dataset-macro accuracy (\%). Right: 15-bin top-label $\mathrm{ECE}_{15}$ (\%), where lower is better. Datasets pool formal records once before equal-dataset averaging; bold marks per-angle best; D10 Box uses $\mathrm{IoU}\ge 0.5$.}
\label{fig:domains}
\end{figure*}
All five systems answer the same non-Box records: three System One models---hosted Jev~1.14 and open-weight JEV-27B-VL and d3-27B---and two general VLMs, Qwen3.8-27B and Gemma~4~31B~IT; Jev~1.14, Qwen3.8-27B, and Gemma~4~31B~IT also complete all Box queries. System One models return interface probabilities, with JEV-27B-VL adapting ordinal grades via Choice, whereas general VLMs normalize conditional likelihoods over valid candidate codes, yielding unified candidate probabilities while noting that the readout mechanism itself may influence comparisons. Native option sets are preserved and derived items use frozen distractors under each model's documented preprocessor with thinking disabled; see Appendix~\ref{sec:execution}.

\subsection{Reliability and Diagnostic Protocols}
\paragraph{Probability calibration and LODO transfer.}
Probability scoring uses an independent Test split where available and source-group cross-validation otherwise, fitting temperatures only on frozen development folds \citep{guo2017calibration,geifman2017selective}. Leave-one-dataset-out LODO transfer then simulates deployment on an unseen task: for each of the 42 probability-eligible datasets, temperature and a global or family-level risk cutoff are fitted on donor datasets alone with the target dataset and shared sources removed, and applied to the target as detailed in Appendix~\ref{sec:lodo}.

\paragraph{\textsc{CrossRender} presentation and native evidence pairs.}
\textsc{CrossRender} is a controlled $2\times 2$ panel crossing text or image stems with text or image options across TT, IT, TI, and II while holding source images, option order, and labels fixed. All five models answer 5,000 originals from 27 datasets in every condition; we report the source-disjoint held-out Test split with cluster-bootstrap intervals and Holm correction in Appendix~\ref{sec:presentation_app}. The 1,640 evidence pairs from MultiBBQ \citep{chen2026multibbq} keep image, question, and options fixed while adding disambiguating text context, with Unknown valid when context is insufficient; see Appendix~\ref{sec:evidencepairs}.

%% file: section/experiments.tex
\section{Experiments}
\label{sec:experiments}

\input{tables/overall}

\begin{figure}[t]
\centering
\includegraphics[width=\columnwidth]{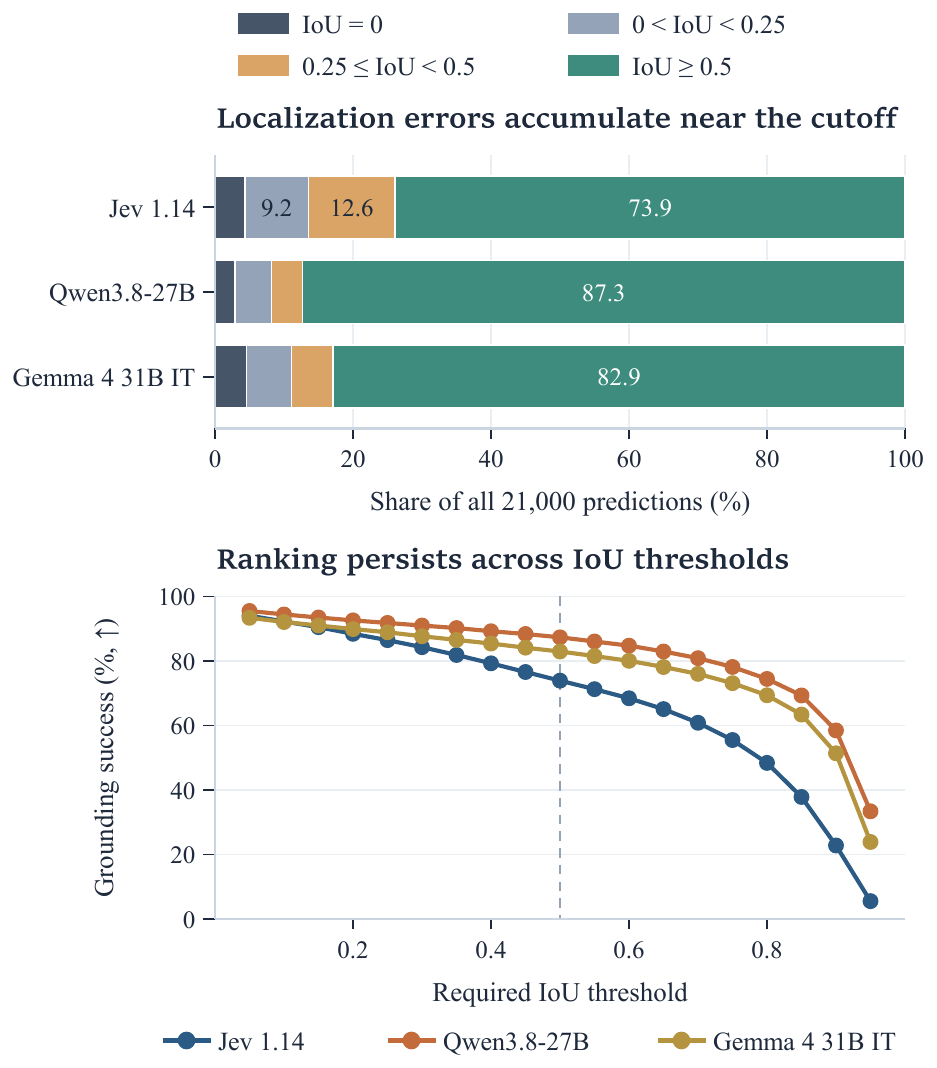}
\caption{Direct Box grounding profile across 21,000 RefCOCO/+/g expressions (D10). Top: full IoU distribution with malformed/zero-area boxes at zero. Bottom: success across IoU thresholds over all queries.}
\label{fig:box}
\end{figure}

\subsection{Main Results}
\label{sec:mainresults}
\paragraph{Interface-level capability spectrum.}
Table~\ref{tab:overall}a shows that System One models lead discrete selection and verification overall. Jev~1.14 ranks first on both Choice and Noul, d3-27B and JEV-27B-VL follow closely, and both general VLMs trail by several points; over all 42 non-Box datasets, Jev~1.14 reaches 80.14\% macro accuracy. The lead is broad across slices: d3-27B beats Qwen on 46 of 56 non-Box slices, as shown in Appendix~\ref{sec:fullresults}. Ordinal and spatial interfaces shift the advantage: Gemma holds a small lead on ordinal exact-grade accuracy, and among the three models evaluated on direct Box grounding, both general VLMs outperform Jev~1.14, with Qwen at 87.34\% and Gemma at 82.94\% versus 73.90\% for Jev~1.14.

\paragraph{Capability and calibration across angles.}
Figure~\ref{fig:domains} places dataset-macro accuracy next to raw $\mathrm{ECE}_{15}$ for D1--D12. Across most discrete perception, reading, and reasoning angles, the System One models combine strong accuracy with low calibration error, led by Jev~1.14 at 4.50\% macro $\mathrm{ECE}_{15}$, though individual angle winners vary across models and metrics. Qwen is moderately overconfident, while Gemma is overconfident almost everywhere at 20.98\% $\mathrm{ECE}_{15}$, placing most of its answers above confidence $0.90$. The System One lead is widest on dense reading and visual structure in D3 OCR and D4 charts, where these models also achieve their lowest $\mathrm{ECE}_{15}$ and outperform both general VLMs on charts by 7.4--10.9 pts. Near-ceiling DocVQA Choice scores reflect closed-set distractors rather than open transcription.

\paragraph{Probability quality and temperature scaling.}
Table~\ref{tab:overall}b,c scores candidate probabilities on the independent Choice Test. No model wins every metric: Jev~1.14 has the lowest raw Brier and ECE, while d3-27B leads NLL and AURC. Dev temperature scaling helps most where raw probabilities are worst, cutting Gemma's NLL from 1.7283 to 0.6321, yet it raises Jev~1.14's ECE even as its NLL improves. Optimizing a proper scoring rule therefore does not guarantee better calibration or a better error ordering, so Table~\ref{tab:all_probability} reports all three separately.

\begin{figure}[t]
\centering
\includegraphics[width=\columnwidth]{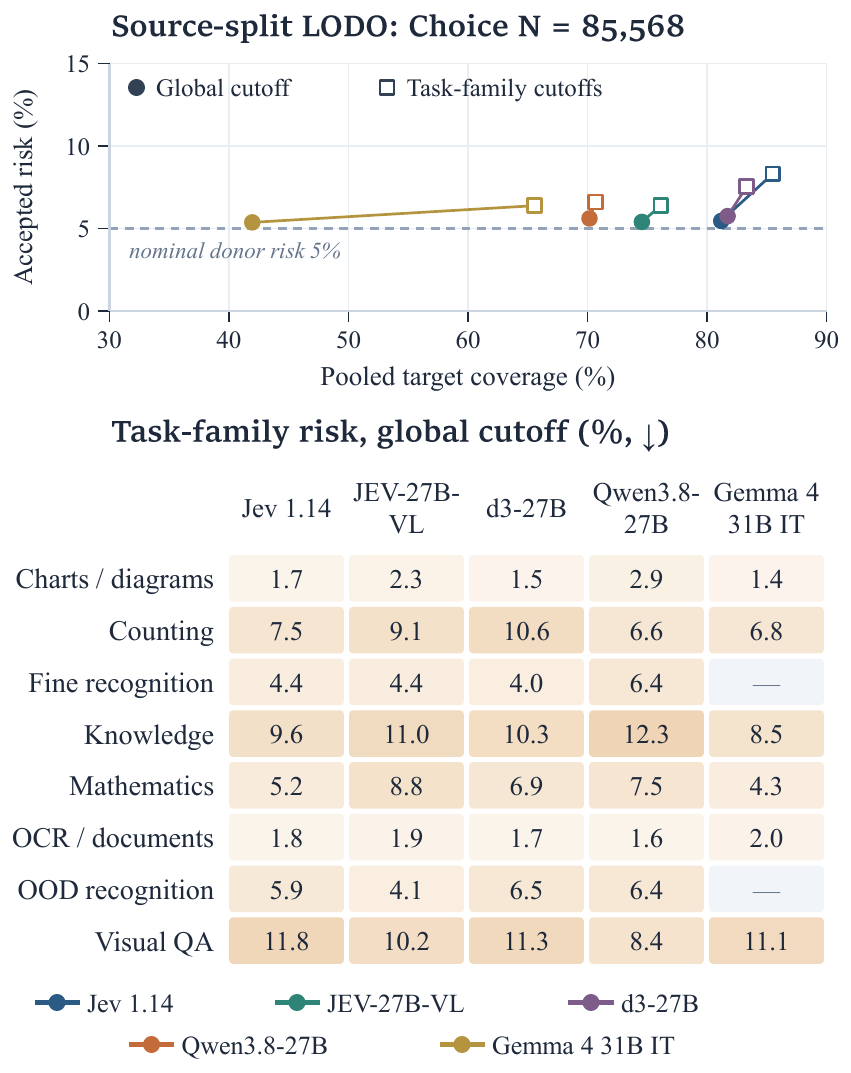}
\caption{Source-disjoint LODO risk transfer at 5\% donor risk on the source-split Choice pool over 25 datasets and $N=85,568$ records. Top: pooled target coverage vs.\ accepted risk for global and task-family cutoffs with 95\% cluster intervals. Bottom: task-family accepted risk (\%) under the global cutoff. Source-fold LODO results over $N=34,410$ records appear in Table~\ref{tab:lodo}b and Figure~\ref{fig:lodo_app}.}
\label{fig:lodo}
\end{figure}

\paragraph{Direct spatial grounding in D10.}
Among the three models with direct Box readout, Figure~\ref{fig:box} shows that Qwen and Gemma outperform Jev~1.14 across IoU thresholds, reaching 87.34\% and 82.94\% versus 73.90\% at $\mathrm{IoU}\ge 0.5$. Most of Jev~1.14's additional errors are near misses rather than zero-overlap boxes: it places 2,639 boxes in $\mathrm{IoU}\in[0.25,0.5)$ versus Qwen's 936 and Gemma's 1,252 in Table~\ref{tab:box}. The gap thus lies mainly in localization precision around the threshold, showing that direct coordinate grounding forms a distinct regime from closed-choice selection.

\subsection{Diagnostic Analysis and Discussion}
\label{sec:diagnostics}
\subsubsection{Four Behavioral Regimes Across D1--D12}
Reading Figure~\ref{fig:domains} by how accuracy and calibration behave together groups D1--D12 into four regimes.
\textbf{First, near-ceiling perception and reading across D1--D4 and D6:} System One models are both accurate and well calibrated, and their high-confidence answers are almost always right.
\textbf{Second, reasoning and multi-image cliffs across D5, D7, and D8:} accuracy falls for every model from D5 knowledge and multilingual exams through D7 cross-view reasoning to D8 abstract induction, where four-way VisuLogic sits near chance. Confidence behavior separates the models here: Gemma's calibration collapses to 51.61\% $\mathrm{ECE}_{15}$ on D8, whereas Jev~1.14 also becomes overconfident on average but keeps its $\mathrm{Risk}@90$ at 2.64\% by avoiding confidence $\ge 0.90$ when uncertain.
\textbf{Third, output-interface shifts across D9 and D10:} on ordinal grading, Gemma's small accuracy edge comes with almost three times Jev~1.14's calibration error; among the three direct-Box models, Qwen and Gemma reverse Jev~1.14's discrete-task lead.
\textbf{Fourth, societal evidence and safety trade-offs across D11 and D12:} D11 tests evidence-sensitive social QA on MultiBBQ and demographic QA on FairLens, where overall accuracy reflects task correctness rather than group fairness. D12 spans meme and toxicity QA on M-QUEST alongside hazard classification on MMSafeAware and MTMCS-Bench, measuring hazard recognition rather than generative refusal. On D12, JEV-27B-VL and Jev~1.14 lead macro accuracy, yet overall accuracy hides class-imbalance trade-offs in Table~\ref{tab:safety}: against the majority-unsafe MMSafeAware baseline of 70.54\% accuracy and 50.00\% balanced accuracy, Jev~1.14 leads both at 68.62\% and 57.55\% with 84.49\% unsafe versus 30.61\% safe recall, whereas d3-27B reaches 58.36\% and 56.47\% by balancing 51.87\% safe and 61.07\% unsafe recall. Gemma's D11 lead likewise hides an abstention trade-off resolved below.

\subsubsection{\textsc{CrossRender}: Visual Presentation Decouples Bias from Risk}
\begin{figure}[t]
\centering
\includegraphics[width=\columnwidth]{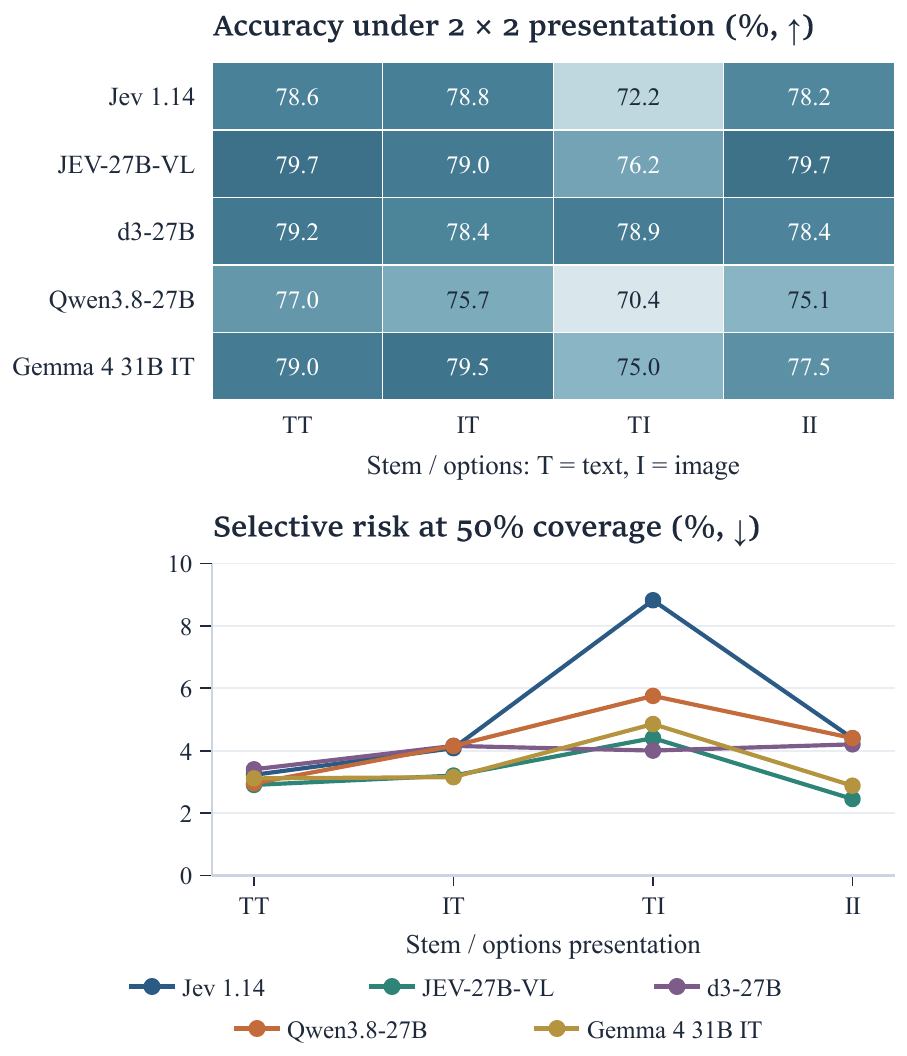}
\caption{\textsc{CrossRender} $2\times 2$ presentation experiment over 3,996 Test originals in 3,737 groups and 100,000 responses. Top: accuracy (\%) under TT, IT, TI, and II, where the two letters give stem and option rendering (T = text, I = image). Bottom: selective risk at 50\% coverage $R_{50}$ (\%). Companion interaction and bias shifts and Unknown-gold MultiBBQ commitments appear in Figure~\ref{fig:presentation_app}.}
\label{fig:presentation}
\end{figure}
Holding item content fixed across stem and option rendering, Figure~\ref{fig:presentation} shows a non-monotonic pattern: rendering \emph{only} the options in TI hurts most, while rendering both stem and options in II largely restores accuracy. Jev~1.14 and Qwen both lose about six points under TI and recover under II, yielding significant positive interactions $I_A>0$ in Table~\ref{tab:presentation}, whereas d3-27B barely moves. Because II largely restores accuracy, the TI drop demonstrates a strong stem-by-option modality interaction; isolating the exact mechanism across cross-modal alignment, OCR legibility, and layout requires further ablations, as discussed in Appendix~\ref{sec:ablations}.

The larger effect is on selective risk. Under TI, Jev~1.14's mean signed confidence bias stays nearly flat at $+0.07$ pts, yet its fixed-coverage risk $R_{50}$ nearly triples from 3.23\% to 8.82\%. Since mean signed bias can cancel opposite errors, its stability does not imply intact error ranking: error-ordering capacity degrades sharply at flat mean signed bias. On Unknown-gold MultiBBQ items in Figure~\ref{fig:presentation_app}, TI raises unsupported answers by over 25 pts for Jev~1.14 and Qwen, whereas d3-27B stays stable.

\subsubsection{Native Evidence Pairs Unpack D11 Trade-Offs}
\input{tables/lodo}
MultiBBQ evidence pairs hold image, question, and options fixed and add disambiguating text in the sufficient state $E^+$, tracking in Table~\ref{tab:evidence_paired}a how models shift between abstaining via Unknown and committing. Jev~1.14 answers both states correctly most often at 87.80\% pair accuracy, combining 89.15\% Unknown recall $U^-$ when textual context is insufficient with 97.87\% accuracy $A^+$ when disambiguating text is present, followed closely by JEV-27B-VL at 86.04\% pair accuracy. The pairs also explain Gemma's D11 aggregate lead: Gemma rarely guesses without context, with unsupported commitment rate $O^- = 4.70\%$, but over-abstains at $U^+ = 10.37\%$ when text suffices. d3-27B shows the opposite trade-off, reaching the highest accuracy of 98.60\% with sufficient text at the cost of 15.00\% unsupported commitments under insufficient text, while Qwen commits without support on 22.87\% of $E^-$ items. These opposite failure modes cancel in single-state accuracy and emerge only in paired evaluation across matched insufficient and sufficient contexts, as detailed in Appendix~\ref{sec:evidencepairs}.

\subsubsection{Cross-Dataset Risk Transfer Under LODO}
A deployed selective-prediction policy is tuned on donor data and applied to unseen tasks. Under our source-disjoint LODO protocol, Table~\ref{tab:lodo}b and Figure~\ref{fig:lodo} show that donor-tuned 5\% risk cutoffs often exceed target risk on unseen datasets, varying across protocols and task families. At a 5\% donor target on Choice, Jev~1.14, JEV-27B-VL, and d3-27B retain 74.55--81.72\% target coverage at 5.40--5.75\% pooled risk under source-split LODO, whereas Gemma accepts only 41.95\% of items to reach 5.37\% risk. Under source-fold LODO in Figure~\ref{fig:lodo_app}, target coverage drops to 11.07--29.30\% and accepted risk rises to 5.98--10.92\% on Choice and up to 29.07\% on Noul. Even when pooled risk stays near 5\%, the bottom panel of Figure~\ref{fig:lodo} shows wide task-family spread: charts and OCR remain below 2\% accepted risk, whereas knowledge and general VQA reach 10--12\%. Fitting task-family cutoffs raises target coverage but also increases pooled risk, showing that donor thresholds under this protocol do not uniformly bound risk across task families.

%% file: tables/overall.tex
\begin{table}[t]
\centering
\renewcommand{\arraystretch}{0.94}
\scriptsize
\setlength{\tabcolsep}{3.1pt}
\begin{tabular}{l rrrr}
\toprule
\sectrow{5}{\textbf{(a) Interface dataset-macro accuracy (\%$\uparrow$; 45 datasets, $N\!=\!192,975$)}} \\
\textbf{Model} & \shortstack{Choice$\uparrow$\\{\tiny (35; 128,079)}} & \shortstack{Noul$\uparrow$\\{\tiny (18; 32,779)}} & \shortstack{Ordinal$\uparrow$\\{\tiny (3; 11,117)}} & \shortstack{Box$\uparrow$\\{\tiny (3; 21,000)}} \\
\midrule
Jev 1.14 & \bestcell{80.49} & \bestcell{90.61} & 73.07 & 73.90 \\
JEV-27B-VL & 78.96 & \secondcell{89.65} & 72.43 & --- \\
d3-27B & \secondcell{80.26} & 89.54 & 73.30 & --- \\
Qwen3.8-27B & 75.13 & 86.35 & \secondcell{73.67} & \bestcell{87.34} \\
Gemma 4 31B IT & 76.15 & 84.18 & \bestcell{75.30} & \secondcell{82.94} \\
\midrule
\sectrow{5}{\textbf{(b) Choice Test raw probability quality ($\downarrow$; 25 datasets, $N\!=\!72,811$)}} \\
\textbf{Model} & NLL$\downarrow$ & Brier$\downarrow$ & ECE$\downarrow$ & AURC$\downarrow$ \\
\midrule
Jev 1.14 & 0.4758 & \bestcell{0.1589} & \bestcell{0.0261} & \secondcell{0.0324} \\
JEV-27B-VL & \secondcell{0.4453} & 0.1938 & 0.0379 & 0.0410 \\
d3-27B & \bestcell{0.3642} & \secondcell{0.1632} & \secondcell{0.0327} & \bestcell{0.0322} \\
Qwen3.8-27B & 0.5769 & 0.2396 & 0.0478 & 0.0563 \\
Gemma 4 31B IT & 1.7283 & 0.3154 & 0.1491 & 0.0735 \\
\midrule
\sectrow{5}{\textbf{(c) Choice Test temperature-scaled probability quality (dev-TS, $\downarrow$)}} \\
Jev 1.14 & 0.4448 & \secondcell{0.1726} & 0.0632 & \secondcell{0.0329} \\
JEV-27B-VL & \secondcell{0.4418} & 0.1923 & \secondcell{0.0329} & 0.0410 \\
d3-27B & \bestcell{0.3619} & \bestcell{0.1625} & \bestcell{0.0283} & \bestcell{0.0322} \\
Qwen3.8-27B & 0.5498 & 0.2358 & 0.0398 & 0.0561 \\
Gemma 4 31B IT & 0.6321 & 0.2557 & 0.0786 & 0.0667 \\
\bottomrule
\end{tabular}
\caption{Main evaluation summary. \textbf{(a)}~Dataset-macro accuracy (\%$\uparrow$) across the four typed interfaces (slice count and $N$ in parentheses; Ordinal reports exact-grade accuracy, Box reports $\mathrm{IoU}\ge 0.5$). \textbf{(b,c)}~Five-model probability quality on independent Choice Test (25 datasets, $N=72,811$; dataset-macro fractions, lower is better) before and after frozen dev temperature scaling ($\mathrm{ECE}$ is 15-bin top-label ECE). Best is \textbf{bolded} (shaded), second \underline{underlined}.}
\label{tab:overall}
\label{tab:probability}
\end{table}

%% file: tables/lodo.tex
\begin{table}[!tb]
\centering
\renewcommand{\arraystretch}{0.90}
\scriptsize
\setlength{\tabcolsep}{2.2pt}
\begin{tabular}{l rr rr r}
\toprule
\sectrow{6}{\textbf{(a) Native MultiBBQ evidence pairs (1,640 pairs, 410 scenes; \%)}} \\
& \multicolumn{2}{c}{Insufficient ($E^-$)} & \multicolumn{2}{c}{Sufficient ($E^+$)} & Pair \\
\cmidrule(lr){2-3}\cmidrule(lr){4-5}\cmidrule(l){6-6}
\textbf{Model} & $U^-\!\uparrow$ & $O^-\!\downarrow$ & $A^+\!\uparrow$ & $U^+\!\downarrow$ & Both$\uparrow$ \\
\midrule
Jev 1.14 & \secondcell{89.15} & \secondcell{10.85} & \secondcell{97.87} & \secondcell{0.85} & \bestcell{87.80} \\
JEV-27B-VL & 88.05 & 11.95 & 97.50 & 1.46 & \secondcell{86.04} \\
d3-27B & 85.00 & 15.00 & \bestcell{98.60} & \bestcell{0.79} & 83.78 \\
Qwen3.8-27B & 77.13 & 22.87 & 94.15 & 2.13 & 72.87 \\
Gemma 4 31B & \bestcell{95.30} & \bestcell{4.70} & 88.78 & 10.37 & 84.45 \\
\bottomrule
\end{tabular}

\vspace{3pt}

\begin{tabular}{l rr rr rr}
\toprule
\sectrow{7}{\textbf{(b) Source-disjoint LODO at 5\% donor risk (42 datasets; \%)}} \\
& \multicolumn{2}{c}{Choice} & \multicolumn{2}{c}{Noul} & \multicolumn{2}{c}{Score} \\
\cmidrule(lr){2-3}\cmidrule(lr){4-5}\cmidrule(l){6-7}
\textbf{Model} & C$\uparrow$ & R$\downarrow$ & C$\uparrow$ & R$\downarrow$ & C$\uparrow$ & R$\downarrow$ \\
\midrule
\sectrow{7}{Source-split LODO: 31 datasets, $N=122,109$} \\
Jev 1.14 & \secondcell{81.19} & 5.46 & 85.16 & 5.41 & 12.29 & 8.13 \\
JEV-27B-VL & 74.55 & \secondcell{5.40} & \secondcell{85.74} & 5.26 & 29.44 & \bestcell{4.68} \\
d3-27B & \bestcell{81.72} & 5.75 & \bestcell{91.22} & 5.27 & \bestcell{44.89} & 6.52 \\
Qwen3.8-27B & 70.17 & 5.61 & 72.98 & \bestcell{5.10} & 15.04 & 7.25 \\
Gemma 4 31B & 41.95 & \bestcell{5.37} & 53.15 & 5.88 & \secondcell{36.82} & \secondcell{5.45} \\
\addlinespace[2pt]
\sectrow{7}{Source-fold LODO: 11 datasets, $N=41,751$} \\
Jev 1.14 & 18.51 & 10.82 & 18.10 & \secondcell{7.22} & --- & --- \\
JEV-27B-VL & \bestcell{29.30} & \bestcell{5.98} & \bestcell{52.65} & \bestcell{6.26} & --- & --- \\
d3-27B & \secondcell{23.26} & 8.15 & 29.72 & 14.39 & --- & --- \\
Qwen3.8-27B & 11.07 & \secondcell{7.93} & \secondcell{32.72} & 7.41 & --- & --- \\
Gemma 4 31B & 13.35 & 10.92 & 14.48 & 29.07 & --- & --- \\
\bottomrule
\end{tabular}
\caption{Decision-reliability validation diagnostics. \textbf{(a)}~Native MultiBBQ evidence pairs ($U^-\!\uparrow$: Unknown recall without context; $O^-\!=1-U^-\!\downarrow$: unsupported judgments; $A^+\!\uparrow$: accuracy with context; $U^+\!\downarrow$: over-abstention; Both$\uparrow$: pair-level accuracy). \textbf{(b)}~Source-disjoint LODO transfer under global scaled thresholds at 5\% donor risk (pooled coverage C\%$\uparrow$ and accepted risk R\%$\downarrow$); dashes denote unevaluated cells. Best is \textbf{bolded} (shaded), second \underline{underlined}.}
\label{tab:evidence_paired}
\label{tab:lodo}
\end{table}

%% file: section/conclusion.tex
\section{Conclusion}
\label{sec:conclusion}
We presented \textbf{OmniDecisionBench}, a benchmark that evaluates the capability and decision reliability of multimodal System One models and general VLMs across 45 datasets, four typed interfaces, and a 5-family $\times$ 12-angle taxonomy, complemented by controlled diagnostics of visual presentation, textual evidence, and target-dataset shift. The two model classes have complementary strengths: System One models lead most discrete perception, reading, and reasoning tasks with better calibration, whereas general VLMs lead ordinal grading and spatial grounding. Multi-image and abstract reasoning remain hard for every model, where confidence rather than accuracy separates them. Reliability also shifts with the input: changes in visual layout, textual context, and target dataset can raise selective risk even when mean confidence bias stays flat. These findings suggest that model choice should follow the task angle and interface, and that confidence thresholds should be re-validated whenever the input distribution changes.

%% file: section/limitations.tex
\section*{Limitations}

First, OmniDecisionBench evaluates single-turn, typed decisions over static images. It does not yet cover video, audio, and multi-turn interaction, where decisions can depend on temporal context or accumulated dialogue state; extending our capability and probability-reliability protocols to these settings is left for future work.

Second, most constituent datasets are in English, and only a small number include multilingual content. Our conclusions therefore mainly describe English-language decision-making, and whether the observed capability and calibration patterns hold across languages and scripts remains to be tested.

Finally, we assess decision quality and probability reliability but do not measure inference latency, throughput, or per-decision cost. Because these factors shape practical deployment, we plan to report them alongside accuracy and calibration in future releases of the benchmark.

%% file: section/ethics.tex
\section*{Ethics Statement}
This research relies exclusively on publicly available multimodal benchmarks and their existing annotations; no human subjects were recruited, and no personal identities or private attributes were inferred. Societal bias and safety benchmarks (MultiBBQ, FairLens, M-QUEST, MMSafeAware, and MTMCS-Bench) are used strictly to diagnose model sensitivity to visual evidence sufficiency and risk-recognition skewness; dataset labels reflect benchmark test specifications rather than normative endorsements or demographic generalizations. All visual assets remain governed by their original benchmark licenses and terms of use. Accordingly, reproducibility artifacts release task manifests, transformation scripts, cached predictions, and evaluation statistics without redistributing restricted source images.

%% file: section/appendix.tex
\section{MECE 5-Family \texorpdfstring{$\times$}{x} 12-Angle Dataset Taxonomy}
\label{sec:taxonomy_app}
Table~\ref{tab:taxonomy45} details the mutually exclusive, collectively exhaustive (MECE) assignment of all 45 public benchmarks (192,975 formal records across 59 dataset--interface slices) to the five capability families (F1--F5) and twelve primary evaluation angles (D1--D12) introduced in Section~\ref{sec:taxonomy}. Every dataset is assigned to a single primary family and angle so it is counted once in macro averages (without implying that underlying skills such as OCR and reasoning are conceptually disjoint); datasets spanning both Choice and Noul slices pool all non-Box formal records once before equal-dataset averaging within their angle.
\input{tables/taxonomy_45}

\section{Complete Dataset Results}
\label{sec:fullresults}
Table~\ref{tab:full_1} reports slice-level accuracy across all 59 formal dataset--interface slices. All five models are evaluated on the same 56 non-Box slices (42 datasets, 171,975 records); Jev 1.14, Qwen3.8-27B, and Gemma 4 31B IT additionally cover the three Box slices. The 307-item M-QUEST toxicity subset discussed in the main text is nested within the 609 M-QUEST records rather than double-counted. In this inventory, MMMU-Pro pools its official configurations, whereas Section~\ref{sec:diagnostics} uses the separate controlled \textsc{CrossRender} panel.
\input{tables/full_1}

\input{section/dataset_citations}

\section{Metric Definitions}
\label{sec:metrics}
A decision instance $x_i=(V_i,q_i,\mathcal{C}_i,\tau_i)$ consists of visual inputs $V_i$ (single or multiple images, or an explicitly flagged text-only control in ScienceQA), task query $q_i$, typed output space $\mathcal{C}_i$, and interface $\tau_i\in\{\text{Choice},\text{Noul},\text{Score},\text{Box}\}$. On discrete interfaces, model $m$ provides a candidate probability distribution $p_m(\cdot\mid x_i)$ over $\mathcal{C}_i$ (returned directly by the interface for System One models and computed as normalized conditional likelihoods over valid candidate codes for general VLMs) with top-label confidence $c_i=\max_{k\in\mathcal{C}_i}p_m(k\mid x_i)$. Let $z_i=\mathbf{1}[\hat y_i\in G_i]$ indicate whether prediction $\hat y_i$ belongs to the accepted gold answer set $G_i$. Across $D$ source benchmarks, unweighted dataset-macro accuracy is
\begin{equation}
A_{\mathrm{macro}}=\frac{1}{D}\sum_{d=1}^{D}\frac{1}{n_d}\sum_{i\in d}z_i,
\end{equation}
whereas record-pooled accuracy weights every instance equally; we never interchange these two summaries. Within each typed interface, our primary aggregate averages participating datasets equally, and equal-slice comparisons across interfaces are explicitly labeled as such.

\paragraph{Probabilistic scoring and calibration.}
For multi-label gold sets $G_i$, set negative log-likelihood is $-\log\sum_{k\in G_i}p_{ik}$. Multiclass Brier score evaluates the full candidate vector on uniquely labeled items, reducing to $(p_i-y_i)^2$ for binary Noul propositions. Partitioning top confidence into 15 equal-width bins $B_b$, expected calibration error is
\begin{equation}
\mathrm{ECE}=\sum_{b=1}^{15}\frac{|B_b|}{n}
\left|\frac{\sum_{i\in B_b}c_i}{|B_b|}-\frac{\sum_{i\in B_b}z_i}{|B_b|}\right|,
\end{equation}
with binary Noul calibration substituting the positive-class probability and binary target. For $K$-level ordinal tasks with cumulative predicted distribution $F_{ik}=\sum_{j\leq k}p_{ij}$ and cumulative target $H_{ik}=\mathbf{1}[y_i\leq k]$, the normalized ranked probability score is
\begin{equation}
\mathrm{RPS}=\frac{1}{n(K-1)}\sum_i\sum_{k=1}^{K-1}(F_{ik}-H_{ik})^2.
\end{equation}
Expected grade ($\sum_{k=1}^K k p_{ik}$) and exact-grade mode accuracy ($\arg\max_k p_{ik}$) capture distinct aspects of ordinal performance and are kept separate.

\paragraph{Selective prediction and confidence bias.}
At threshold $t$ with accepted set $S_t=\{i:c_i\geq t\}$, coverage is $|S_t|/n$ and selective risk is $\sum_{i\in S_t}(1-z_i)/|S_t|$. $R_{50}$ is the selective risk at the threshold reaching 50\% coverage, and $\mathrm{Risk}@90$ is the selective risk of $S_{0.90}$. AURC averages cumulative selective risk across the confidence ranking using exact expectation over tied confidence groups, while threshold policies accept or reject complete tie groups atomically. Unlike non-negative ECE, signed confidence bias $B = n^{-1}\sum_i(c_i-z_i)$ preserves error direction, and presentation-induced bias drift $\Delta B=B_{\mathrm{TI}}-B_{\mathrm{TT}}$ measures the signed shift between matched visual formats. For \textsc{CrossRender} condition accuracies $A_{\mathrm{TT}},A_{\mathrm{IT}},A_{\mathrm{TI}},A_{\mathrm{II}}$, the accuracy interaction is $I_A=A_{\mathrm{II}}-A_{\mathrm{IT}}-A_{\mathrm{TI}}+A_{\mathrm{TT}}$.

\paragraph{Post-hoc scaling and spatial grounding.}
From cached candidate probabilities, temperature scaling computes
\begin{equation}
p'_{ik}(T)=\frac{\exp(\log\max(p_{ik},\epsilon)/T)}{\sum_j\exp(\log\max(p_{ij},\epsilon)/T)},
\end{equation}
with floor $\epsilon=10^{-12}$ in all reliability analyses; any $T>0$ preserves top-1 accuracy. Bounding-box IoU divides intersection area by union area in original image coordinates, counting malformed responses and degenerate zero-area boxes as incorrect ($0$ IoU) rather than dropping them from the denominator.

\section{Five-Model Probability Results}
All five models are evaluated under the independent-Test and source-group cross-validation protocols in Table~\ref{tab:all_probability}. The independent Test contains 101,894 visual records from 31 datasets; the source-group evaluation contains 41,751 records from 11 datasets. Dataset-macro probability scores retain their own protocol and output mode. Top-label ECE evaluates confidence in the predicted answer, whereas native Noul event ECE evaluates the positive proposition and is retained separately in the numerical archive. API probabilities are preserved as returned, including zeros; NLL uses a $10^{-12}$ floor, with $10^{-6}$ sensitivity supplied for \textsc{CrossRender}.
\input{tables/all_probability}

\section{Source-Disjoint Leave-One-Dataset-Out Evaluation}
\label{sec:lodo}
\begin{figure*}[t]
\centering
\includegraphics[width=0.86\textwidth]{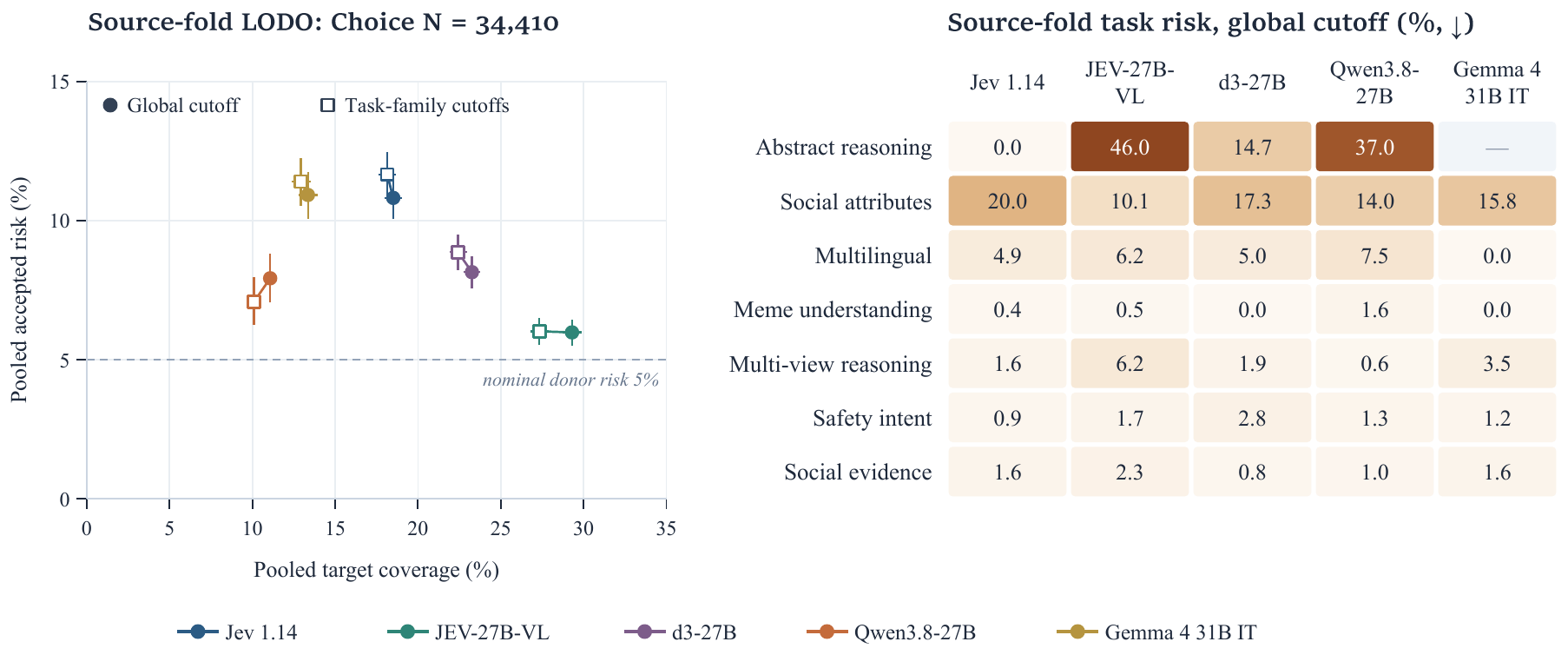}
\caption{Companion source-fold LODO risk transfer at 5\% donor risk across the 10 five-fold Choice datasets ($N=34,410$). Left: pooled target coverage vs.\ accepted risk for global and task-family cutoffs (95\% source-cluster intervals). Right: task-family accepted risk (\%) under the global cutoff.}
\label{fig:lodo_app}
\end{figure*}
The transfer population contains 42 unique probability-eligible datasets and 163,860 records. The 122,109 source-split records retain archived fit/risk/Test source roles; the 41,751 source-fold records retain five folds. For target fold $k$, donor risk uses fold $k+1$ and donor temperature uses the other three folds; the entire target dataset and all shared source components are removed from every donor role. Fitting uses same-mode donors and dataset-equal NLL. No target labels are used for fitting. Temperatures are chosen on an 81-point grid in $[0.05,20]$ minimizing dataset-equal NLL. Global or family cutoffs target 1\% or 5\% donor risk and require at least 100 accepted items and 50 source components. Family thresholds use the same global temperature and a family-specific donor-risk cutoff, falling back globally only for insufficient donor support. Intervals use 2,000 source-cluster resamples.

All 1\% and 5\% operating points, accepted/error counts, task exposures, and probability scores are retained in the accompanying CSV source data. Figures~\ref{fig:lodo} and~\ref{fig:lodo_app} use record-pooled risk and the matching pooled intervals. Dataset-macro risk is a different estimand and is never combined with pooled-count intervals. Reported 95\% intervals condition on the frozen fits; refitting donors or deriving a simultaneous distribution-free guarantee is outside this computation.

\section{Native Evidence Pairs and Prediction Transitions}
\label{sec:evidencepairs}
Of 4,912 sampled MultiBBQ records, 3,280 form 1,640 complete visual-language evidence pairs across 410 scenes, giving 16,400 cached answers over five models; inference uses 8,192 scene resamples with Holm correction over 60 tests. The remaining 1,632 are not forced into unmatched pairs. Each source scene's generators and question polarities remain together in bootstrap sampling. Pair accuracy is the proportion correct in both conditions, not the mean of the two separate accuracies. Unknown denotes the task's uncertainty option; it is neither a model-output failure nor an external rejection.

The sufficient condition retains the ambiguous context and adds the original unmasked disambiguating text, including the released demographic descriptions, as in the frozen execution adapter. Images, question text, and ordered candidates are identical within each pair. The comparison tests use of additional textual evidence in a multimodal task; it does not identify a response to visual evidence removal. A full request-reconstruction audit verifies 3,280 API hashes; the released-model source-task hash and cached d3 prompt strings independently preserve the execution conditions.
\input{tables/evidence_confusion}

\section{\textsc{CrossRender} Presentation Protocol}
\label{sec:presentation_app}
\begin{figure*}[t]
\centering
\includegraphics[width=0.86\textwidth]{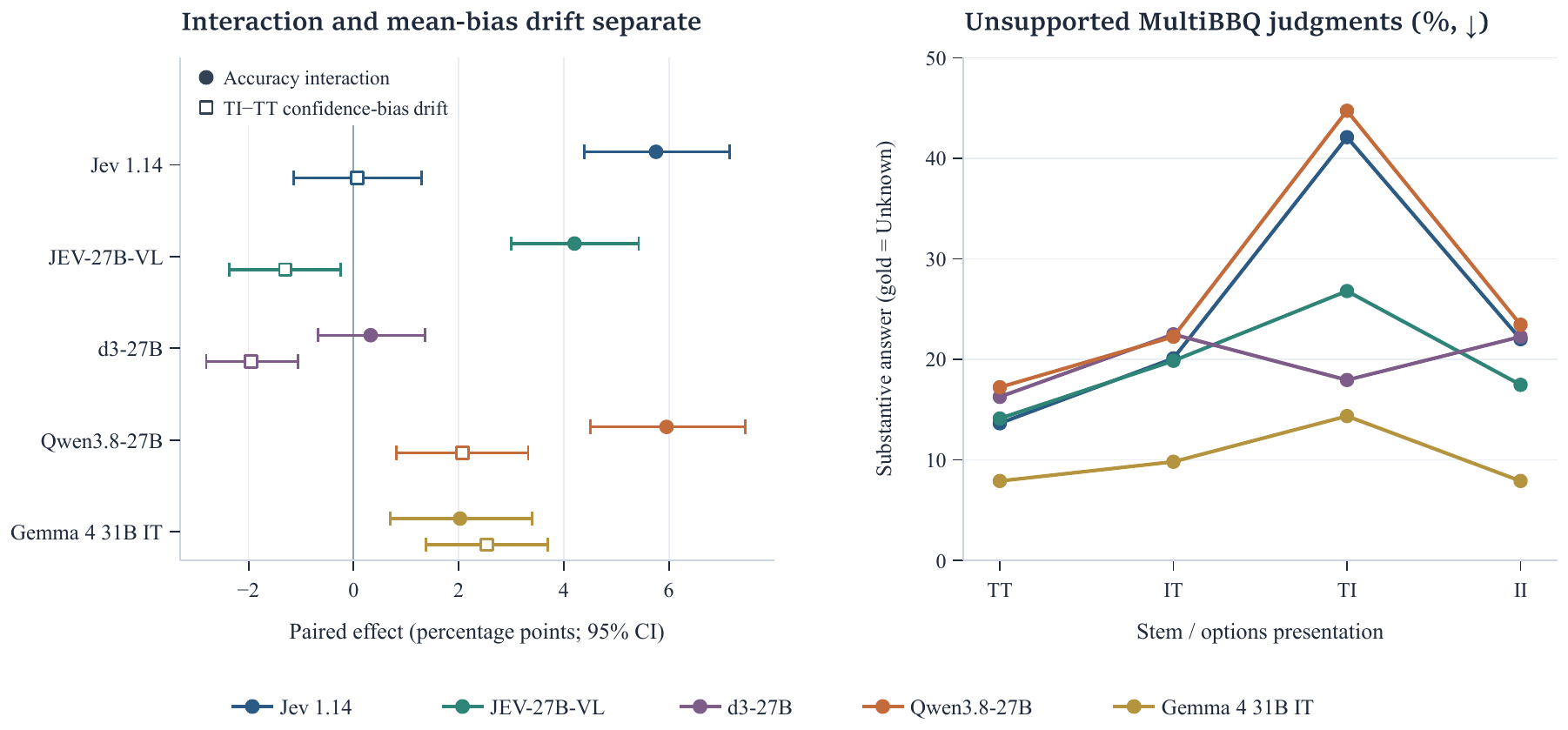}
\caption{Companion diagnostics for the \textsc{CrossRender} $2\times 2$ presentation panel (3,996 Test originals). Left: accuracy interaction $I_A$ versus TI--TT signed confidence-bias drift $\Delta B$ (percentage points; 95\% cluster-bootstrap intervals, 8,192 resamples). Right: unsupported substantive commitments (\%) on 418 Unknown-gold MultiBBQ Test originals across TT, IT, TI, and II.}
\label{fig:presentation_app}
\end{figure*}
Five models complete all four conditions of the same 5,000 originals, giving 100,000 total responses rather than 100,000 independent questions. The panel covers 27 datasets and nine domains, with 1,004 development originals and 3,996 Test originals (3,737 source components; 8,192 cluster resamples). Rendered option semantics are withheld from text prompts in the image-option conditions. All variants and models share source splits. The primary accuracy family contains 105 paired contrasts; raw signed-bias drift uses 45 contrasts and raw $R_{50}$ uses 30. Holm corrections are applied within these archived families. Marginal percentile intervals do not imply simultaneous coverage. Tests use a plus-one correction with minimum attainable $p=1/8193$.

The confidence-bias quantity $B=\bar c-A$ is signed and can cancel heterogeneous errors. Its stability is therefore tested alongside NLL, Brier, ECE, AURC, fixed-coverage risk, and evidence-specific answer selection. A temperature fitted on TT development is fixed for all four Test presentations; it is not refitted after inspecting TI or II. Because every original appears in all four conditions, this controlled comparison isolates presentation effects that an observational standard-versus-screenshot split (e.g., MMMU-Pro's official configurations) cannot separate from item differences.

\section{Input and Execution Reproducibility}
\label{sec:execution}
Table~\ref{tab:checkpoints} lists exact model identifiers, archived revision prefixes, and consolidated diagnostic summaries ($2\times 2$ presentation effects, MMSafeAware safety recall, and direct Box IoU bins). Smaller datasets are retained in full; larger ones use a deterministic 7,000-record cap. Local checkpoints run in BF16. All deterministic sampling steps use seed 42. On the societal and harder visual datasets for Qwen3.8-27B and Gemma 4 31B IT, frozen image preprocessing converts EXIF-oriented RGB inputs to JPEG (quality 95), caps the longest edge at 1,568 pixels, and pads edges below 280 pixels; JEV-27B-VL and d3-27B pass original image bytes directly to their native processors. Cross-model comparisons thus share identical source visual evidence under each release's documented preprocessing pipeline.
\input{tables/additional_numerical}

Every task manifest records dataset, split, source question and image IDs, prompt text, option semantics, accepted gold labels, output interface, and transformation version, cleanly separating native multiple-choice items, exact programmatic mappings, and derived distractor sets. Text-only ScienceQA items are flagged explicitly as non-visual controls, and Kaleidoscope is restricted to image-required items.

All local runs execute in BF16 with thinking disabled, renormalizing likelihoods over valid option codes for closed general-purpose VLM readouts---a constrained conditional distribution distinct from unconstrained sequence likelihood. Resumable execution preserves completed responses while separating transport-layer failures from invalid model outputs, and grounding evaluations use Qwen3.8-27B's corrected native coordinate run uniformly rather than mixing coordinate conventions.

Finally, our table-generation audit verifies SHA-256 hashes, sample denominators, and merged cloud grounding retries across every table. Full-suite tables incorporate the merged cloud retries, and RefCOCO IoU histograms are recomputed directly from those merged grades so that every reported Box figure matches the final evaluation totals.

\section{Controls and Mechanism Questions}
\label{sec:ablations}
The \textsc{CrossRender} stem-by-option presentation experiment and native evidence-sufficiency pairs are the controlled interventions in this study. Each changes one input factor at the level of the end-to-end system; neither isolates an individual visual encoder, OCR stage, training objective, or probability head. Shared detector-generated candidate regions would compare referent selection against direct coordinates without generating proposals from test gold boxes; proposal coverage must accompany end-to-end success. Within-question candidate-set scaling and independently reviewed visual-evidence edits address different mechanisms and are natural extensions of the protocol.

%% file: tables/taxonomy_45.tex
\begin{table*}[t]
\centering
\scriptsize
\setlength{\tabcolsep}{2.7pt}
\begin{tabular}{llllrr rrrrr}
\toprule
& & & & & & \multicolumn{5}{c}{Dataset accuracy (\%)$\uparrow$} \\
\cmidrule(l){7-11}
\textbf{Fam.} & \textbf{Angle} & \textbf{Dataset} & Mode(s) & $N$ & Comp. & Jev 1.14 & JEV-27B & d3-27B & Qwen3.8 & Gemma 4 \\
\midrule
F1 & D1 Fine/OOD & Flowers102 & Ch & 6,149 & 6,147 & \bestcell{86.5} & 76.0 & \secondcell{79.5} & 77.3 & 78.9 \\
F1 & D1 Fine/OOD & Food101 & Ch & 7,000 & 6,999 & \secondcell{90.1} & 82.9 & 86.5 & 84.2 & \bestcell{91.7} \\
F1 & D1 Fine/OOD & ImageNet-A & Ch & 7,000 & 6,996 & \secondcell{77.4} & 74.5 & 75.8 & 72.1 & \bestcell{79.3} \\
F1 & D1 Fine/OOD & ImageNet-R & Ch & 7,000 & 6,992 & \bestcell{89.3} & 84.8 & 86.2 & 83.8 & \secondcell{88.6} \\
F1 & D2 Gen.\ VQA & VizWiz & Ch/No & 48 & 48 & \bestcell{97.9} & \secondcell{95.8} & \secondcell{95.8} & 91.7 & \secondcell{95.8} \\
F1 & D2 Gen.\ VQA & MMBench & Ch & 7,000 & 912 & \bestcell{93.0} & 92.1 & \secondcell{92.8} & 90.0 & 90.7 \\
F1 & D2 Gen.\ VQA & MME & No & 2,374 & 1,177 & \bestcell{93.3} & 89.3 & \secondcell{91.2} & 87.3 & 86.7 \\
F1 & D2 Gen.\ VQA & SEED-Bench & Ch & 7,000 & 3,937 & \secondcell{81.2} & 81.0 & \bestcell{81.4} & 78.0 & 79.4 \\
F1 & D2 Gen.\ VQA & RealWorldQA & Ch/No & 725 & 722 & \secondcell{80.7} & 80.1 & \bestcell{83.3} & 79.9 & 72.8 \\
F1 & D2 Gen.\ VQA & GQA & Ch/No & 5,361 & 2,661 & 83.5 & \secondcell{83.9} & \bestcell{85.5} & 81.4 & 79.9 \\
\midrule
F2 & D3 Doc/OCR & DocVQA & Ch/No & 3,461 & 1,171 & 99.7 & \bestcell{99.9} & \bestcell{99.9} & 99.5 & 99.3 \\
F2 & D3 Doc/OCR & OCRBench & Ch/No & 1,000 & 930 & 99.3 & \bestcell{99.7} & \bestcell{99.7} & 98.4 & 98.7 \\
F2 & D3 Doc/OCR & TextVQA & Ch/No & 1,816 & 1,560 & 98.6 & \secondcell{98.9} & \bestcell{99.2} & 98.3 & 97.9 \\
F2 & D3 Doc/OCR & OCR-VQA & Ch/No & 7,000 & 6,101 & \secondcell{90.9} & 90.2 & \bestcell{92.3} & 90.4 & 89.2 \\
F2 & D3 Doc/OCR & InfographicVQA & Ch/No & 1,588 & 432 & \secondcell{89.6} & 88.5 & \bestcell{92.8} & 80.2 & 82.1 \\
F2 & D4 Charts & DVQA & Ch/No & 7,000 & 6,772 & 97.4 & \secondcell{97.6} & \bestcell{99.1} & 95.2 & 92.3 \\
F2 & D4 Charts & ChartQA & Ch/No & 2,024 & 1,374 & \secondcell{95.8} & 94.8 & \bestcell{96.5} & 88.6 & 90.7 \\
F2 & D4 Charts & PlotQA & Ch/No & 6,253 & 5,547 & \secondcell{91.3} & 88.3 & \bestcell{94.8} & 77.1 & 67.0 \\
F2 & D4 Charts & AI2D & Ch & 3,088 & 814 & \bestcell{90.5} & 88.4 & \secondcell{90.2} & 84.5 & 86.9 \\
\midrule
F3 & D5 Exams/ML & ScienceQA & Ch & 4,241 & 4,016 & \secondcell{97.1} & 95.8 & \bestcell{97.3} & 94.4 & 94.9 \\
F3 & D5 Exams/ML & A-OKVQA & Ch & 1,145 & 1,122 & \bestcell{90.8} & 88.8 & \secondcell{89.7} & 86.8 & \secondcell{89.7} \\
F3 & D5 Exams/ML & MMStar & Ch & 1,500 & 1,430 & \secondcell{73.4} & 71.6 & \bestcell{74.5} & 64.8 & 68.7 \\
F3 & D5 Exams/ML & MMMU-Pro & Ch & 5,190 & 1,606 & \bestcell{61.9} & 53.0 & \secondcell{61.1} & 46.9 & 53.2 \\
F3 & D5 Exams/ML & Kaleidoscope & Ch & 7,000 & 6,302 & \secondcell{57.3} & 54.4 & \bestcell{58.6} & 47.8 & 54.7 \\
F3 & D6 Spatial/Cnt & CLEVR & Ch/No & 7,000 & 5,699 & \bestcell{97.0} & 92.9 & \secondcell{96.3} & 85.0 & 66.4 \\
F3 & D6 Spatial/Cnt & NLVR2 & No & 5,852 & 926 & 93.3 & \secondcell{94.6} & \bestcell{95.4} & 92.1 & 91.1 \\
F3 & D6 Spatial/Cnt & VSR & No & 3,292 & 2,627 & \bestcell{86.8} & 86.1 & \secondcell{86.3} & 85.9 & 84.4 \\
F3 & D6 Spatial/Cnt & TallyQA & Ch & 7,000 & 6,263 & \secondcell{62.6} & 61.9 & \bestcell{64.3} & 55.1 & 59.3 \\
F3 & D7 Multi-Img & MVH-Bench & Ch/No & 4,320 & 626 & 82.3 & \secondcell{82.8} & \bestcell{83.1} & 78.8 & 78.3 \\
F3 & D7 Multi-Img & MMSI-Bench & Ch & 1,000 & 756 & 35.1 & \secondcell{36.5} & \bestcell{40.2} & 32.2 & 31.6 \\
F3 & D8 Math/Ind. & MathVista & Ch & 1,000 & 972 & \bestcell{84.7} & 77.9 & \secondcell{82.9} & 67.5 & 66.9 \\
F3 & D8 Math/Ind. & VisReason & Ch & 457 & 456 & \bestcell{31.5} & 29.5 & \secondcell{30.6} & 29.1 & 27.1 \\
F3 & D8 Math/Ind. & VisRes-Bench & Ch & 7,000 & 6,652 & 30.8 & \secondcell{31.4} & \bestcell{32.8} & 27.7 & 30.1 \\
F3 & D8 Math/Ind. & VisuLogic & Ch & 1,000 & 1,000 & 26.5 & 27.3 & \secondcell{29.5} & \bestcell{29.6} & 26.0 \\
\midrule
F4 & D9 Ordinal & WIDER-FACE Occlusion & Sc & 7,000 & 2,131 & 66.3 & \bestcell{68.7} & 67.3 & \secondcell{68.3} & 66.7 \\
F4 & D9 Ordinal & Crisis Damage & Sc & 3,865 & 3,850 & 76.3 & 78.0 & \bestcell{79.2} & 77.3 & \secondcell{78.2} \\
F4 & D9 Ordinal & BRACOL Severity & Sc & 252 & 252 & \secondcell{76.6} & 70.6 & 73.4 & 75.4 & \bestcell{81.0} \\
F4 & D10 Box & RefCOCO & Bx & 7,000 & 2,580 & 74.7 & --- & --- & \bestcell{88.7} & \secondcell{85.0} \\
F4 & D10 Box & RefCOCO+ & Bx & 7,000 & 2,574 & 67.8 & --- & --- & \bestcell{84.2} & \secondcell{76.6} \\
F4 & D10 Box & RefCOCOg & Bx & 7,000 & 3,221 & 79.2 & --- & --- & \bestcell{89.1} & \secondcell{87.3} \\
\midrule
F5 & D11 Evidence & MultiBBQ & Ch & 4,912 & 410 & \secondcell{89.5} & 88.4 & 88.5 & 81.2 & \bestcell{91.7} \\
F5 & D11 Evidence & FairLens & Ch & 7,000 & 1,324 & 73.6 & \secondcell{82.6} & 67.8 & 77.6 & \bestcell{88.9} \\
F5 & D12 Safety & M-QUEST & Ch & 609 & 307 & \secondcell{90.1} & \bestcell{91.5} & 88.3 & 81.6 & 85.6 \\
F5 & D12 Safety & MMSafeAware & No & 1,453 & 445 & \bestcell{68.6} & \secondcell{67.3} & 58.4 & 60.0 & 52.0 \\
F5 & D12 Safety & MTMCS-Bench & Ch/No & 7,000 & 721 & 87.8 & \bestcell{89.2} & \secondcell{88.7} & 88.0 & 86.8 \\
\bottomrule
\end{tabular}
\caption{Complete 45-dataset MECE taxonomy inventory across five capability families (F1--F5) and twelve mutually exclusive evaluation angles (D1--D12), totaling 192,975 formal records. Modes: Ch=Choice, No=Noul, Sc=Score, Bx=Box. Dataset accuracy (\%$\uparrow$) pools all formal records of each dataset; Comp.\ reports unique source components. Best per row is \textbf{bolded} (shaded) and second-best \underline{underlined}.}
\label{tab:taxonomy45}
\end{table*}

%% file: tables/full_1.tex
\begin{table*}[t]
\centering
\renewcommand{\arraystretch}{0.93}
\scriptsize
\setlength{\tabcolsep}{1.75pt}
\begin{tabular}{@{}llr rrrrr@{\hspace{10pt}}llr rrrrr@{}}
\toprule
& & & \multicolumn{5}{c}{Slice accuracy (\%)$\uparrow$} & & & & \multicolumn{5}{c}{Slice accuracy (\%)$\uparrow$} \\
\cmidrule(lr){4-8}\cmidrule(l){12-16}
\textbf{Dataset} & Type & $N$ & \shortstack{Jev\\1.14} & \shortstack{JEV-\\27B} & \shortstack{d3-\\27B} & \shortstack{Qwen\\3.8} & \shortstack{Gemma\\4 31B} & \textbf{Dataset} & Type & $N$ & \shortstack{Jev\\1.14} & \shortstack{JEV-\\27B} & \shortstack{d3-\\27B} & \shortstack{Qwen\\3.8} & \shortstack{Gemma\\4 31B} \\
\midrule
Flowers-102 & C & 6,149 & \bestcell{86.50} & 76.01 & \secondcell{79.48} & 77.35 & 78.94 & DVQA & N & 2,573 & 95.18 & \bestcell{99.34} & \secondcell{99.07} & 98.68 & 96.58 \\
Food-101 & C & 7,000 & \secondcell{90.13} & 82.91 & 86.49 & 84.19 & \bestcell{91.70} & PlotQA & C & 5,051 & \secondcell{91.07} & 87.33 & \bestcell{95.19} & 75.07 & 62.70 \\
MMBench & C & 7,000 & \bestcell{92.96} & 92.11 & \secondcell{92.76} & 90.03 & 90.71 & PlotQA & N & 1,202 & \secondcell{92.35} & \secondcell{92.35} & \bestcell{93.26} & 85.61 & 85.27 \\
RealWorldQA & C & 565 & \secondcell{80.00} & 79.82 & \bestcell{83.36} & 79.47 & 74.34 & MathVista & C & 1,000 & \bestcell{84.70} & 77.90 & \secondcell{82.90} & 67.50 & 66.90 \\
RealWorldQA & N & 160 & \bestcell{83.12} & 81.25 & \bestcell{83.12} & 81.25 & 67.50 & MMMU-Pro & C & 5,190 & \bestcell{61.87} & 53.01 & \secondcell{61.12} & 46.90 & 53.20 \\
GQA & C & 1,480 & \bestcell{92.03} & \secondcell{91.62} & 91.55 & 88.31 & 89.46 & MMStar & C & 1,500 & \secondcell{73.40} & 71.60 & \bestcell{74.47} & 64.80 & 68.73 \\
GQA & N & 3,881 & 80.19 & \secondcell{80.98} & \bestcell{83.12} & 78.77 & 76.19 & MME & N & 2,374 & \bestcell{93.26} & 89.34 & \secondcell{91.15} & 87.32 & 86.73 \\
VizWiz & C & 20 & \bestcell{100.00} & \bestcell{100.00} & 95.00 & 95.00 & \bestcell{100.00} & SEED-Bench & C & 7,000 & \secondcell{81.17} & 81.00 & \bestcell{81.40} & 78.00 & 79.40 \\
VizWiz & N & 28 & \bestcell{96.43} & 92.86 & \bestcell{96.43} & 89.29 & 92.86 & CLEVR & C & 2,800 & \secondcell{95.43} & 90.79 & \bestcell{97.00} & 83.04 & 61.61 \\
MVH-Bench & C & 1,440 & \secondcell{84.03} & \bestcell{85.07} & 82.85 & 80.21 & 79.79 & CLEVR & N & 4,200 & \bestcell{98.00} & 94.24 & \secondcell{95.88} & 86.26 & 69.57 \\
MVH-Bench & N & 2,880 & 81.46 & \secondcell{81.67} & \bestcell{83.23} & 78.16 & 77.60 & TallyQA & C & 7,000 & \secondcell{62.56} & 61.86 & \bestcell{64.33} & 55.07 & 59.29 \\
DocVQA & C & 3,434 & 99.74 & \secondcell{99.85} & \bestcell{99.94} & 99.53 & 99.27 & NLVR2 & N & 5,852 & 93.30 & \secondcell{94.62} & \bestcell{95.44} & 92.11 & 91.11 \\
DocVQA & N & 27 & \bestcell{100.00} & \bestcell{100.00} & \bestcell{100.00} & \bestcell{100.00} & \bestcell{100.00} & VSR & N & 3,292 & \bestcell{86.76} & 86.09 & \secondcell{86.30} & 85.87 & 84.36 \\
TextVQA & C & 1,755 & 98.58 & \secondcell{99.09} & \bestcell{99.26} & 98.46 & 98.01 & RefCOCO & B & 7,000 & 74.70 & --- & --- & \bestcell{88.71} & \secondcell{84.96} \\
TextVQA & N & 61 & \bestcell{98.36} & 93.44 & \secondcell{96.72} & 93.44 & 95.08 & RefCOCO+ & B & 7,000 & 67.83 & --- & --- & \bestcell{84.21} & \secondcell{76.57} \\
OCR-VQA & C & 5,315 & \bestcell{93.45} & 91.98 & \secondcell{93.04} & 91.42 & 91.46 & RefCOCOg & B & 7,000 & 79.19 & --- & --- & \bestcell{89.10} & \secondcell{87.29} \\
OCR-VQA & N & 1,685 & 82.97 & 84.39 & \bestcell{90.15} & \secondcell{87.18} & 81.96 & VisRes Bench & C & 7,000 & 30.79 & \secondcell{31.39} & \bestcell{32.77} & 27.73 & 30.10 \\
OCRBench & C & 995 & 99.30 & \secondcell{99.70} & \bestcell{99.80} & 98.49 & 98.79 & VisReason & C & 457 & \bestcell{31.51} & 29.54 & \secondcell{30.63} & 29.10 & 27.13 \\
OCRBench & N & 5 & \bestcell{100.00} & \bestcell{100.00} & 80.00 & 80.00 & 80.00 & MMSI-Bench & C & 1,000 & 35.10 & \secondcell{36.50} & \bestcell{40.20} & 32.20 & 31.60 \\
InfographicVQA & C & 1,587 & \secondcell{89.60} & 88.47 & \bestcell{92.75} & 80.21 & 82.04 & VisuLogic & C & 1,000 & 26.50 & 27.30 & \secondcell{29.50} & \bestcell{29.60} & 26.00 \\
InfographicVQA & N & 1 & \bestcell{100.00} & \bestcell{100.00} & \bestcell{100.00} & \bestcell{100.00} & \bestcell{100.00} & MultiBBQ & C & 4,912 & \secondcell{89.52} & 88.42 & 88.48 & 81.25 & \bestcell{91.69} \\
WIDER FACE & O & 7,000 & 66.30 & \bestcell{68.66} & 67.26 & \secondcell{68.29} & 66.73 & FairLens & C & 7,000 & 73.64 & \secondcell{82.60} & 67.84 & 77.60 & \bestcell{88.94} \\
BRACOL & O & 252 & \secondcell{76.59} & 70.63 & 73.41 & 75.40 & \bestcell{80.95} & Kaleidoscope & C & 7,000 & \secondcell{57.27} & 54.36 & \bestcell{58.60} & 47.84 & 54.73 \\
Crisis & O & 3,865 & 76.33 & 78.01 & \bestcell{79.22} & 77.34 & \secondcell{78.21} & ImageNet-A & C & 7,000 & \secondcell{77.37} & 74.53 & 75.77 & 72.10 & \bestcell{79.30} \\
A-OKVQA & C & 1,145 & \bestcell{90.83} & 88.82 & \secondcell{89.69} & 86.81 & \secondcell{89.69} & ImageNet-R & C & 7,000 & \bestcell{89.26} & 84.83 & 86.24 & 83.79 & \secondcell{88.60} \\
ScienceQA & C & 4,241 & \secondcell{97.15} & 95.78 & \bestcell{97.26} & 94.39 & 94.93 & M-QUEST & C & 609 & \secondcell{90.15} & \bestcell{91.46} & 88.34 & 81.61 & 85.55 \\
AI2D & C & 3,088 & \bestcell{90.48} & 88.41 & \secondcell{90.16} & 84.49 & 86.95 & MMSafeAware & N & 1,453 & \bestcell{68.62} & \secondcell{67.31} & 58.36 & 60.01 & 52.03 \\
ChartQA & C & 1,927 & \secondcell{95.95} & 95.33 & \bestcell{97.04} & 89.10 & 90.92 & MTMCS-Bench & C & 3,992 & \secondcell{86.60} & \bestcell{87.68} & 84.69 & 85.80 & 83.02 \\
ChartQA & N & 97 & \bestcell{91.75} & 84.54 & 85.57 & 79.38 & \secondcell{86.60} & MTMCS-Bench & N & 3,008 & 89.33 & 91.22 & \bestcell{93.98} & 90.89 & \secondcell{91.76} \\
DVQA & C & 4,427 & \secondcell{98.69} & 96.61 & \bestcell{99.14} & 93.11 & 89.77 & & & & & & & & \\
\bottomrule
\end{tabular}
\caption{Complete slice-level benchmark results across all 59 formal dataset--interface slices (45 datasets, 192,975 records). Values report accuracy (\%$\uparrow$), with Box (B) evaluated at $\mathrm{IoU}\geq 0.5$. Interface codes: C = Choice, N = Noul, O = Ordinal, B = Box. Dashes denote unevaluated cells; best per slice is \textbf{bolded} (shaded) and second-best \underline{underlined}.}
\label{tab:full_1}
\label{tab:full_2}
\label{tab:full_3}
\label{tab:full_4}
\end{table*}

%% file: section/dataset_citations.tex
\subsection{Dataset sources}
\label{app:dataset_sources}
Source benchmarks in the core suite are grouped by capability domain below (extension benchmarks are cited in Section~\ref{sec:related}); exact splits and interface mappings are recorded in the accompanying task manifests.

\paragraph{Fine-grained recognition and distribution shift.}
Food-101 and Flowers-102 evaluate fine-grained visual categorization, complemented by natural adversarial and rendition shifts in ImageNet-A and ImageNet-R \citep{bossard2014food101,nilsback2008flowers102,hendrycks2021imageneta,hendrycks2021imagenetr}.

\paragraph{General visual question answering and broad suites.}
Open-domain and knowledge-intensive VQA draw on GQA, A-OKVQA, VizWiz, and RealWorldQA \citep{hudson2019gqa,schwenk2022aokvqa,gurari2018vizwiz,xai2024realworldqa}, alongside broad multimodal suites including MMBench, MME, MathVista, ScienceQA, and SEED-Bench \citep{liu2023mmbench,fu2023mme,lu2024mathvista,lu2022scienceqa,li2023seedbench}.

\paragraph{Charts, diagrams, and document OCR.}
Quantitative diagram and chart interpretation uses ChartQA, PlotQA, DVQA, and AI2D \citep{masry2022chartqa,methani2020plotqa,kafle2018dvqa,kembhavi2016ai2d}; dense text and document reading uses DocVQA, TextVQA, OCR-VQA, and InfographicVQA \citep{mathew2021docvqa,singh2019textvqa,mishra2019ocrvqa,mathew2022infographicvqa}.

\paragraph{Compositional, counting, and spatial reasoning.}
CLEVR, TallyQA, NLVR2, and VSR test compositional visual logic, object counting, multi-image verification, and spatial relations \citep{johnson2017clevr,acharya2019tallyqa,suhr2019nlvr2,liu2023vsr}.

%% file: tables/all_probability.tex
\begin{table*}[t]
\centering
\renewcommand{\arraystretch}{0.94}
\scriptsize
\setlength{\tabcolsep}{3.4pt}
\begin{tabular}{l rrrr @{\hspace{14pt}} l rrrr}
\toprule
\multicolumn{5}{c}{\textbf{Independent Test Protocol (31 datasets, $N=101,894$)}} & \multicolumn{5}{c}{\textbf{Source-Group Cross-Validation Protocol (11 datasets, $N=41,751$)}} \\
\cmidrule(r){1-5}\cmidrule(l){6-10}
\textbf{Model} & NLL$\downarrow$ & Brier$\downarrow$ & ECE$\downarrow$ & AURC$\downarrow$ & \textbf{Model} & NLL$\downarrow$ & Brier$\downarrow$ & ECE$\downarrow$ & AURC$\downarrow$ \\
\midrule
\sectrow{5}{Choice (25 datasets; $N=72,811$)} & \sectrow{5}{Choice (10 datasets; $N=34,410$)} \\
Jev 1.14 & 0.4758 & \bestcell{0.1589} & \bestcell{0.0261} & \secondcell{0.0324} & Jev 1.14 & 0.9049 & \bestcell{0.4686} & \bestcell{0.0743} & 0.3062 \\
JEV-27B-VL & \secondcell{0.4453} & 0.1938 & 0.0379 & 0.0410 & JEV-27B-VL & \secondcell{0.8778} & \secondcell{0.4691} & 0.0925 & \secondcell{0.2987} \\
d3-27B & \bestcell{0.3642} & \secondcell{0.1632} & \secondcell{0.0327} & \bestcell{0.0322} & d3-27B & \bestcell{0.8770} & 0.4789 & \secondcell{0.0855} & \bestcell{0.2913} \\
Qwen3.8-27B & 0.5769 & 0.2396 & 0.0478 & 0.0563 & Qwen3.8-27B & 1.0630 & 0.5515 & 0.1432 & 0.3392 \\
Gemma 4 31B IT & 1.7283 & 0.3154 & 0.1491 & 0.0735 & Gemma 4 31B IT & 3.1914 & 0.7277 & 0.3491 & 0.3286 \\
\addlinespace[2pt]
\sectrow{5}{Noul (15 datasets; $N=20,255$)} & \sectrow{5}{Noul (3 datasets; $N=7,341$)} \\
Jev 1.14 & \secondcell{0.2191} & \secondcell{0.0619} & 0.0695 & \bestcell{0.0253} & Jev 1.14 & \secondcell{0.4403} & \secondcell{0.1425} & \secondcell{0.0410} & \bestcell{0.1208} \\
JEV-27B-VL & 0.2282 & 0.0669 & \secondcell{0.0490} & 0.0300 & JEV-27B-VL & \bestcell{0.4299} & \bestcell{0.1398} & \bestcell{0.0407} & \secondcell{0.1212} \\
d3-27B & \bestcell{0.1928} & \bestcell{0.0569} & \bestcell{0.0368} & \secondcell{0.0267} & d3-27B & 0.4831 & 0.1529 & 0.0800 & 0.1506 \\
Qwen3.8-27B & 0.2994 & 0.0881 & 0.0624 & 0.0498 & Qwen3.8-27B & 0.4924 & 0.1595 & 0.0626 & 0.1624 \\
Gemma 4 31B IT & 1.6162 & 0.1332 & 0.1352 & 0.0805 & Gemma 4 31B IT & 3.4859 & 0.2543 & 0.2531 & 0.2058 \\
\addlinespace[2pt]
\sectrow{5}{Score (3 datasets; $N=8,828$)} & \multicolumn{5}{l}{\textit{(No source-group CV slices for Score)}} \\
Jev 1.14 & 0.7134 & 0.3644 & 0.0812 & 0.1139 & & & & & \\
JEV-27B-VL & \secondcell{0.6288} & \secondcell{0.3498} & \bestcell{0.0601} & \secondcell{0.1021} & & & & & \\
d3-27B & \bestcell{0.5999} & \bestcell{0.3392} & 0.0813 & \bestcell{0.0893} & & & & & \\
Qwen3.8-27B & 0.6796 & 0.3702 & \secondcell{0.0747} & 0.1171 & & & & & \\
Gemma 4 31B IT & 2.4703 & 0.4668 & 0.2306 & 0.1191 & & & & & \\
\bottomrule
\end{tabular}
\caption{Raw five-model probability metrics (lower is better, $\downarrow$) across both frozen validation protocols (Independent Test on the left; Source-Group Cross-Validation on the right) and output modes. Dataset-macro means are computed separately within each protocol and mode; Noul uses native binary Brier loss and ECE calibrates top-label confidence against correctness. Best per block is \textbf{bolded} (shaded) and second-best \underline{underlined}.}
\label{tab:all_probability}
\end{table*}

%% file: tables/evidence_confusion.tex
\begin{table*}[t]
\centering
\renewcommand{\arraystretch}{0.94}
\scriptsize
\setlength{\tabcolsep}{4.0pt}
\begin{tabular}{l rrr rrr rrr}
\toprule
\sectrow{10}{\textbf{(a) Paired prediction transitions from insufficient ($E^-$) to sufficient ($E^+$) context (1,640 pairs; row sum = 1,640)}} \\
& \multicolumn{3}{c}{Insufficient = U} & \multicolumn{3}{c}{Insufficient = T} & \multicolumn{3}{c}{Insufficient = O} \\
\cmidrule(lr){2-4}\cmidrule(lr){5-7}\cmidrule(l){8-10}
\textbf{Model} & U/U$\downarrow$ & U/T$\uparrow$ & U/O$\downarrow$ & T/U$\downarrow$ & T/T$\downarrow$ & T/O$\downarrow$ & O/U$\downarrow$ & O/T$\downarrow$ & O/O$\downarrow$ \\
\midrule
Jev 1.14 & 12 & \bestcell{1,440} & 10 & 0 & 77 & 0 & 2 & 88 & 11 \\
JEV-27B-VL & 24 & \secondcell{1,411} & 9 & 0 & 90 & 0 & 0 & 98 & 8 \\
d3-27B & 13 & 1,374 & 7 & 0 & 113 & 0 & 0 & 130 & 3 \\
Qwen3.8-27B & 31 & 1,195 & 39 & 2 & 195 & 1 & 2 & 154 & 21 \\
Gemma 4 31B IT & 170 & 1,385 & 8 & 0 & 28 & 0 & 0 & 43 & 6 \\
\bottomrule
\end{tabular}

\vspace{3pt}

\setlength{\tabcolsep}{3.5pt}
\begin{tabular}{ll rrr @{\hspace{12pt}} ll rrr}
\toprule
\sectrow{10}{\textbf{(b) MultiBBQ paired model contrasts from 8,192 source-scene bootstrap draws (seed 20261119; Holm-adjusted over 60 tests)}} \\
\textbf{Contrast} & \textbf{Endpoint} & $\Delta$ (pp) & 95\% CI & $p_{\rm Holm}$ & \textbf{Contrast} & \textbf{Endpoint} & $\Delta$ (pp) & 95\% CI & $p_{\rm Holm}$ \\
\midrule
Jev 1.14 $-$ JEV-27B-VL & Both correct ($\uparrow$) & 1.77 & [0.00, 3.54] & 0.6357 & Jev 1.14 $-$ Qwen3.8-27B & Both correct ($\uparrow$) & 14.94 & [12.26, 17.74] & 0.0073 \\
Jev 1.14 $-$ JEV-27B-VL & Unsupported ($\downarrow$) & $-$1.10 & [$-$2.62, 0.43] & 1.0000 & Jev 1.14 $-$ Qwen3.8-27B & Unsupported ($\downarrow$) & $-$12.01 & [$-$14.51, $-$9.57] & 0.0073 \\
Jev 1.14 $-$ JEV-27B-VL & Over-abstention ($\downarrow$) & $-$0.61 & [$-$1.52, 0.24] & 1.0000 & Jev 1.14 $-$ Qwen3.8-27B & Over-abstention ($\downarrow$) & $-$1.28 & [$-$2.32, $-$0.37] & 0.1930 \\
\cmidrule(r){1-5}\cmidrule(l){6-10}
Jev 1.14 $-$ d3-27B & Both correct ($\uparrow$) & 4.02 & [1.77, 6.28] & 0.0081 & Jev 1.14 $-$ Gemma 4 31B & Both correct ($\uparrow$) & 3.35 & [0.43, 6.28] & 0.3918 \\
Jev 1.14 $-$ d3-27B & Unsupported ($\downarrow$) & $-$4.15 & [$-$6.16, $-$2.20] & 0.0073 & Jev 1.14 $-$ Gemma 4 31B & Unsupported ($\downarrow$) & 6.16 & [4.51, 7.99] & 0.0073 \\
Jev 1.14 $-$ d3-27B & Over-abstention ($\downarrow$) & 0.06 & [$-$0.79, 0.85] & 1.0000 & Jev 1.14 $-$ Gemma 4 31B & Over-abstention ($\downarrow$) & $-$9.51 & [$-$11.95, $-$7.13] & 0.0073 \\
\bottomrule
\end{tabular}
\caption{MultiBBQ native evidence-pair transition matrix and bootstrap inference. \textbf{(a)}~Paired prediction transitions (U=Unknown, T=sufficient-condition gold target, O=other substantive option; only U/T is correct in both states). \textbf{(b)}~Paired bootstrap contrasts (positive $\Delta$ favors Jev~1.14 on Both correct $\uparrow$; negative $\Delta$ favors Jev~1.14 on Unsupported $\downarrow$ and Over-abstention $\downarrow$).}
\label{tab:evidence_confusion}
\label{tab:paired}
\end{table*}

%% file: tables/additional_numerical.tex
\begin{table*}[t]
\centering
\renewcommand{\arraystretch}{0.94}
\scriptsize
\setlength{\tabcolsep}{3.3pt}
\begin{tabular}{l ll rrrr rrr}
\toprule
\sectrow{10}{\textbf{(a) Evaluated Model Checkpoints \& \textsc{CrossRender} $2\times 2$ Presentation Diagnostics (3,996 Test originals; 100,000 total responses)}} \\
& \multicolumn{2}{c}{Checkpoint Provenance} & \multicolumn{4}{c}{$2\times 2$ Accuracy (\%)$\uparrow$} & \multicolumn{3}{c}{Presentation Effects (pp)} \\
\cmidrule(lr){2-3}\cmidrule(lr){4-7}\cmidrule(l){8-10}
\textbf{Model} & \textbf{Evaluated identifier} & \textbf{Revision / release} & TT$\uparrow$ & IT$\uparrow$ & TI$\uparrow$ & II$\uparrow$ & $I_A$ & $\Delta B$ & $\Delta R_{50}\!\downarrow$ \\
\midrule
Jev 1.14 & \texttt{jev-1.14.0} & Hosted release & 78.58 & 78.75 & 72.22 & 78.15 & 5.76 & 0.07 & 5.60 \\
JEV-27B-VL & \texttt{autotrust/JEV-27B-VL} & \texttt{9e53b8a809d2} & \bestcell{79.73} & \secondcell{79.00} & \secondcell{76.23} & \bestcell{79.70} & 4.20 & $-$1.30 & \secondcell{1.50} \\
d3-27B & \texttt{vllm-sr/d3} & \texttt{62f07e54c6c3} & \secondcell{79.20} & 78.38 & \bestcell{78.85} & \secondcell{78.35} & 0.33 & $-$1.95 & \bestcell{0.60} \\
Qwen3.8-27B & \texttt{Qwen/Qwen3.8-27B} & \texttt{6e218df9c238} & 77.03 & 75.70 & 70.42 & 75.05 & 5.96 & 2.07 & 2.80 \\
Gemma 4 31B IT & \texttt{google/gemma-4-31B-it} & \texttt{842da3794eaa} & 78.98 & \bestcell{79.45} & 75.00 & 77.50 & 2.03 & 2.53 & 1.74 \\
\bottomrule
\end{tabular}

\vspace{3pt}

\setlength{\tabcolsep}{2.7pt}
\begin{tabular}{l rrrr @{\hspace{12pt}} l rrrr r}
\toprule
\multicolumn{5}{c}{\textbf{(b) MMSafeAware Binary Safety ($N=1,453$, \%; \texttt{Yes} = safe)}} & \multicolumn{6}{c}{\textbf{(c) Direct Box Grounding by IoU Bin (21,000 RefCOCO/+/g queries)}} \\
\cmidrule(r){1-5}\cmidrule(l){6-11}
\textbf{Model} & Acc.$\uparrow$ & Bal.\ Acc.$\uparrow$ & Safe Rec.$\uparrow$ & Unsafe Rec.$\uparrow$ & \textbf{Model} & $=0\downarrow$ & $(0,.25)\downarrow$ & $[.25,.5)\downarrow$ & $\geq .5\uparrow$ & Succ.\ (\%)$\uparrow$ \\
\midrule
Jev 1.14 & \bestcell{68.62} & \bestcell{57.55} & 30.61 & \secondcell{84.49} & Jev 1.14 & \secondcell{912} & 1,929 & 2,639 & 15,520 & 73.90 \\
JEV-27B-VL & \secondcell{67.31} & 52.95 & 17.99 & \bestcell{87.90} & Qwen3.8-27B & \bestcell{607} & \bestcell{1,115} & \bestcell{936} & \bestcell{18,342} & \bestcell{87.34} \\
d3-27B & 58.36 & \secondcell{56.47} & \secondcell{51.87} & 61.07 & Gemma 4 31B IT & 959 & \secondcell{1,372} & \secondcell{1,252} & \secondcell{17,417} & \secondcell{82.94} \\
Qwen3.8-27B & 60.01 & 52.13 & 32.94 & 71.32 & \multicolumn{6}{l}{\textit{(JEV-27B-VL and d3-27B do not support direct Box readout)}} \\
Gemma 4 31B IT & 52.03 & 53.21 & \bestcell{56.07} & 50.34 & & & & & & \\
\bottomrule
\end{tabular}
\caption{Consolidated reproducibility and diagnostic summary. \textbf{(a)}~Evaluated model identifiers, archived commit prefixes, and \textsc{CrossRender} $2\times 2$ presentation results ($I_A=A_{II}-A_{IT}-A_{TI}+A_{TT}$, $\Delta B=B_{TI}-B_{TT}$, and 50\%-coverage risk change $\Delta R_{50}$ in percentage points). \textbf{(b)}~MMSafeAware binary safety classification (predicting unsafe everywhere yields 70.54\% accuracy and 50.00\% balanced accuracy). \textbf{(c)}~RefCOCO/+/g direct Box grounding counts by IoU bin (malformed outputs count as $\mathrm{IoU}=0$). Best per directional column is \textbf{bolded} (shaded), second \underline{underlined}.}
\label{tab:checkpoints}
\label{tab:presentation}
\label{tab:safety}
\label{tab:box}
\label{tab:coverage}
\label{tab:selected}
\end{table*}

%% file: references.bib
@misc{deusser2026jev,
      title={{Evaluating and Benchmarking the System One Model Jev}}, 
      author={Tobias Deu{\ss}er and Rajkumar Ramamurthy and Lorenz Sparrenberg and Rafet Sifa},
      year={2026},
      eprint={2609.37647},
      archivePrefix={arXiv},
      primaryClass={cs.CL},
      url={https://arxiv.org/abs/2609.37647v2}, 
      note={Version 2, 29 October 2026},
}

@inproceedings{lee2024vhelm,
 author = {Lee, Tony and Tu, Haoqin and Wong, Chi Heem and Zheng, Wenhao and Zhou, Yiyang and Mai, Yifan and Roberts, Josselin Somerville and Yasunaga, Michihiro and Yao, Huaxiu and Xie, Cihang and Liang, Percy},
 booktitle = {Advances in Neural Information Processing Systems},
 doi = {10.52202/079017-4464},
 editor = {A. Globerson and L. Mackey and D. Belgrave and A. Fan and U. Paquet and J. Tomczak and C. Zhang},
 pages = {140632--140666},
 publisher = {Curran Associates, Inc.},
 title = {{VHELM: A Holistic Evaluation of Vision Language Models}},
 url = {https://proceedings.neurips.cc/paper_files/paper/2024/file/fe2fc7dc60b55ccd8886220b40fb1f74-Paper-Datasets_and_Benchmarks_Track.pdf},
 volume = {37},
 year = {2024}
}

@article{liang2023helm,
  title     = {{Holistic Evaluation of Language Models}},
  author    = {Liang, Percy and Bommasani, Rishi and Lee, Tony and Tsipras, Dimitris and Soylu, Dilara and Yasunaga, Michihiro and Zhang, Yian and Narayanan, Deepak and Wu, Yuhuai and Kumar, Ananya and Newman, Benjamin and Yuan, Binhang and Yan, Bobby and Zhang, Ce and Cosgrove, Christian and Manning, Christopher D and Re, Christopher and Acosta-Navas, Diana and Hudson, Drew A. and Zelikman, Eric and Durmus, Esin and Ladhak, Faisal and Rong, Frieda and Ren, Hongyu and Yao, Huaxiu and Wang, Jue and Santhanam, Keshav and Orr, Laurel and Zheng, Lucia and Yuksekgonul, Mert and Suzgun, Mirac and Kim, Nathan and Guha, Neel and Chatterji, Niladri S. and Khattab, Omar and Henderson, Peter and Huang, Qian and Chi, Ryan Andrew and Xie, Sang Michael and Santurkar, Shibani and Ganguli, Surya and Hashimoto, Tatsunori and Icard, Thomas and Zhang, Tianyi and Chaudhary, Vishrav and Wang, William and Li, Xuechen and Mai, Yifan and Zhang, Yuhui and Koreeda, Yuta},
  journal   = {Transactions on Machine Learning Research},
  year      = {2023},
  url       = {https://mlanthology.org/tmlr/2023/liang2023tmlr-holistic/}
}

@InProceedings{guo2017calibration,
  title = 	 {{On Calibration of Modern Neural Networks}},
  author =       {Chuan Guo and Geoff Pleiss and Yu Sun and Kilian Q. Weinberger},
  booktitle = 	 {Proceedings of the 34th International Conference on Machine Learning},
  pages = 	 {1321--1330},
  year = 	 {2017},
  editor = 	 {Precup, Doina and Teh, Yee Whye},
  volume = 	 {70},
  series = 	 {Proceedings of Machine Learning Research},
  month = 	 {06--11 Aug},
  publisher =    {PMLR},
  url = 	 {https://proceedings.mlr.press/v70/guo17a.html},
}

@misc{geifman2017selective,
      title={{Selective Classification for Deep Neural Networks}}, 
      author={Yonatan Geifman and Ran El-Yaniv},
      year={2017},
      eprint={1705.08500},
      archivePrefix={arXiv},
      primaryClass={cs.LG},
      url={https://arxiv.org/abs/1705.08500}, 
}

@misc{yue2024mmmupro,
      title={{MMMU-Pro: A More Robust Multi-discipline Multimodal Understanding Benchmark}}, 
      author={Xiang Yue and Tianyu Zheng and Yuansheng Ni and Yubo Wang and Kai Zhang and Shengbang Tong and Yuxuan Sun and Botao Yu and Ge Zhang and Huan Sun and Yu Su and Wenhu Chen and Graham Neubig},
      year={2025},
      eprint={2409.02813},
      archivePrefix={arXiv},
      primaryClass={cs.CL},
      url={https://arxiv.org/abs/2409.02813}, 
}

@misc{chen2024mmstar,
      title={{Are We on the Right Way for Evaluating Large Vision-Language Models?}}, 
      author={Lin Chen and Jinsong Li and Xiaoyi Dong and Pan Zhang and Yuhang Zang and Zehui Chen and Haodong Duan and Jiaqi Wang and Yu Qiao and Dahua Lin and Feng Zhao},
      year={2024},
      eprint={2403.20330},
      archivePrefix={arXiv},
      primaryClass={cs.CV},
      url={https://arxiv.org/abs/2403.20330}, 
}

@inproceedings{yu2016refcoco, title={{Modeling Context in Referring Expressions}}, ISBN={9783319464756}, ISSN={1611-3349}, url={http://dx.doi.org/10.1007/978-3-319-46475-6_5}, DOI={10.1007/978-3-319-46475-6_5}, booktitle={Computer Vision -- ECCV 2016}, publisher={Springer International Publishing}, author={Yu, Licheng and Poirson, Patrick and Yang, Shan and Berg, Alexander C. and Berg, Tamara L.}, year={2016}, pages={69--85} }

@inproceedings{chen2026multibbq,
  title  =  {{Fairness Failure Modes of Multimodal {LLM}s}},
  author  =  {Chen, Canyu and Cai, Anglin and Nwatu, Joan and Li, Yale and Hullman, Jessica and Mihalcea, Rada and McKeown, Kathleen and Li, Manling},
  booktitle  =  {The Fortieth Annual Conference on Neural Information Processing Systems},
  year  =  {2026},
  url  =  {https://openreview.net/forum?id=myI2BmrxeL}
}

@misc{khazaie2026fairlens,
      title={{FairLens: Benchmarking Fairness in Vision-Language Models for High-Stakes Decision-Making}}, 
      author={Vahid Reza Khazaie and Ahmed Y. Radwan and Shaina Raza},
      year={2026},
      eprint={2609.01691},
      archivePrefix={arXiv},
      primaryClass={cs.CV},
      url={https://arxiv.org/abs/2609.01691}, 
}

@inproceedings{wang2025mmsafeaware,
    title = "Can{'}t See the Forest for the Trees: Benchmarking Multimodal Safety Awareness for Multimodal {LLM}s",
    author = "Wang, Wenxuan  and
      Liu, Xiaoyuan  and
      Gao, Kuiyi  and
      Huang, Jen-tse  and
      Yuan, Youliang  and
      He, Pinjia  and
      Wang, Shuai  and
      Tu, Zhaopeng",
    editor = "Che, Wanxiang  and
      Nabende, Joyce  and
      Shutova, Ekaterina  and
      Pilehvar, Mohammad Taher",
    booktitle = "Proceedings of the 63rd Annual Meeting of the {A}ssociation for {C}omputational {L}inguistics (Volume 1: Long Papers)",
    month = jul,
    year = "2025",
    address = "Vienna, Austria",
    publisher = "Association for Computational Linguistics",
    url = "https://aclanthology.org/2025.acl-long.832/",
    doi = "10.18653/v1/2025.acl-long.832",
    pages = "16993--17006",
    ISBN = "979-8-89176-251-0"
}

@misc{salazar2025kaleidoscope,
      title={{Kaleidoscope: In-language Exams for Massively Multilingual Vision Evaluation}}, 
      author={Israfel Salazar and Manuel Fern{\'{a}}ndez Burda and Shayekh Bin Islam and Arshia Soltani Moakhar and Shivalika Singh and Fabian Farestam and Angelika Romanou and Danylo Boiko and Dipika Khullar and Mike Zhang and Dominik Krzemi{\'{n}}ski and Jekaterina Novikova and Lu{\'{i}}sa Shimabucoro and Joseph Marvin Imperial and Rishabh Maheshwary and Sharad Duwal and Alfonso Amayuelas and Swati Rajwal and Jebish Purbey and Ahmed Ruby and Nicholas Popovi{\v{c}} and Marek Suppa and Azmine Toushik Wasi and Ram Mohan Rao Kadiyala and Olga Tsymboi and Maksim Kostritsya and Bardia Soltani Moakhar and Gabriel da Costa Merlin and Ot{\'{a}}vio Ferracioli Coletti and Maral Jabbari Shiviari and MohammadAmin farahani fard and Silvia Fernandez and Mar{\'{i}}a Grandury and Dmitry Abulkhanov and Drishti Sharma and Andre Guarnier De Mitri and Leticia Bossatto Marchezi and Setayesh Heydari and Johan Obando-Ceron and Nazar Kohut and Beyza Ermis and Desmond Elliott and Enzo Ferrante and Sara Hooker and Marzieh Fadaee},
      year={2025},
      eprint={2504.07072},
      archivePrefix={arXiv},
      primaryClass={cs.CL},
      url={https://arxiv.org/abs/2504.07072}, 
}

@misc{xu2025visulogic,
      title={{VisuLogic: A Benchmark for Evaluating Visual Reasoning in Multi-modal Large Language Models}}, 
      author={Weiye Xu and Jiahao Wang and Weiyun Wang and Zhe Chen and Wengang Zhou and Aijun Yang and Lewei Lu and Houqiang Li and Xiaohua Wang and Xizhou Zhu and Wenhai Wang and Jifeng Dai and Jinguo Zhu},
      year={2025},
      eprint={2504.15279},
      archivePrefix={arXiv},
      primaryClass={cs.CV},
      url={https://arxiv.org/abs/2504.15279}, 
}

@misc{tortei2025visres,
      title={{VisRes Bench: On Evaluating the Visual Reasoning Capabilities of VLMs}}, 
      author={Brigitta Malagurski T{\"{o}}rtei and Yasser Dahou and Ngoc Dung Huynh and Wamiq Reyaz Para and Ph{\'{u}}c H. L{\^{e}} Khac and Ankit Singh and Sofian Chaybouti and Sanath Narayan},
      year={2025},
      eprint={2512.21194},
      archivePrefix={arXiv},
      primaryClass={cs.CV},
      url={https://arxiv.org/abs/2512.21194}, 
}

@misc{yang2025mmsi,
      title={{MMSI-Bench: A Benchmark for Multi-Image Spatial Intelligence}}, 
      author={Sihan Yang and Runsen Xu and Yiman Xie and Sizhe Yang and Mo Li and Jingli Lin and Chenming Zhu and Xiaochen Chen and Haodong Duan and Xiangyu Yue and Dahua Lin and Tai Wang and Jiangmiao Pang},
      year={2026},
      eprint={2505.23764},
      archivePrefix={arXiv},
      primaryClass={cs.CV},
      url={https://arxiv.org/abs/2505.23764}, 
}

@inproceedings{guo2026visreason,
    title = "Can {MLLM}s Reason Beyond Language? {V}is{R}eason: A Comprehensive Benchmark for Vision-Centric Reasoning",
    author = "Guo, Longteng  and
      Wang, Yifan  and
      Huo, Pengkang  and
      Chen, Tailai  and
      Wu, Yuze  and
      Liu, Jing  and
      Zhu, Xinxin",
    editor = "Liakata, Maria  and
      Moreira, Viviane P.  and
      Zhang, Jiajun  and
      Jurgens, David",
    booktitle = "Findings of the {A}ssociation for {C}omputational {L}inguistics: {ACL} 2026",
    month = jul,
    year = "2026",
    address = "San Diego, California, United States",
    publisher = "Association for Computational Linguistics",
    url = "https://aclanthology.org/2026.findings-acl.1996/",
    doi = "10.18653/v1/2026.findings-acl.1996",
    pages = "40149--40192",
    ISBN = "979-8-89176-395-1"
}

@misc{park2026mvh,
      title={{Revealing Multi-View Hallucination in Large Vision-Language Models}}, 
      author={Wooje Park and Insu Lee and Soohyun Kim and Jaeyun Jang and Minyoung Noh and Kyuhong Shim and Byonghyo Shim},
      year={2026},
      eprint={2603.23934},
      archivePrefix={arXiv},
      primaryClass={cs.CV},
      url={https://arxiv.org/abs/2603.23934}, 
}

@misc{degiorgis2026mquest,
      title={{M-QUEST -- Meme Question-Understanding Evaluation on Semantics and Toxicity}}, 
      author={Stefano De Giorgis and Ting-Chih Chen and Filip Ilievski},
      year={2026},
      eprint={2603.03315},
      archivePrefix={arXiv},
      primaryClass={cs.CL},
      url={https://arxiv.org/abs/2603.03315}, 
}

@misc{liu2026mtmcs,
      title={{MTMCS-Bench: Evaluating Contextual Safety of Multimodal Large Language Models in Multi-Turn Dialogues}}, 
      author={Zheyuan Liu and Dongwhi Kim and Yixin Wan and Xiangchi Yuan and Zhaoxuan Tan and Fengran Mo and Meng Jiang},
      year={2026},
      eprint={2601.06757},
      archivePrefix={arXiv},
      primaryClass={cs.CL},
      url={https://arxiv.org/abs/2601.06757}, 
}

@article{liu2024ocrbench,
    title={{OCRBench: on the hidden mystery of OCR in large multimodal models}},
    volume={67},
    ISSN={1869-1919},
    url={http://dx.doi.org/10.1007/s11432-024-4235-6},
    DOI={10.1007/s11432-024-4235-6},
    number={12},
    journal={Science China Information Sciences},
    publisher={Springer Science and Business Media LLC},
    author={Liu, Yuliang and Li, Zhang and Huang, Mingxin and Yang, Biao and Yu, Wenwen and Li, Chunyuan and Yin, Xu-Cheng and Liu, Cheng-Lin and Jin, Lianwen and Bai, Xiang},
    year={2024},
    month=dec }

@InProceedings{yang2016widerface,

author = {Yang, Shuo and Luo, Ping and Loy, Chen-Change and Tang, Xiaoou},

title = {{WIDER FACE: A Face Detection Benchmark}},

booktitle = {Proceedings of the IEEE Conference on Computer Vision and Pattern Recognition (CVPR)},

month = {June},

year = {2016}

}

@inproceedings{alam2020crisis, title={{Deep Learning Benchmarks and Datasets for Social Media Image Classification for Disaster Response}}, url={http://dx.doi.org/10.1109/ASONAM49781.2020.9381294}, DOI={10.1109/asonam49781.2020.9381294}, booktitle={2020 IEEE/ACM International Conference on Advances in Social Networks Analysis and Mining (ASONAM)}, publisher={IEEE}, author={Alam, Firoj and Ofli, Ferda and Imran, Muhammad and Alam, Tanvirul and Qazi, Umair}, year={2020}, month=Dec, pages={151--158} }

@article{esgario2020bracol, title={{Deep learning for classification and severity estimation of coffee leaf biotic stress}}, volume={169}, ISSN={0168-1699}, url={http://dx.doi.org/10.1016/j.compag.2019.105162}, DOI={10.1016/j.compag.2019.105162}, journal={Computers and Electronics in Agriculture}, publisher={Elsevier BV}, author={Esgario, Jos{\'{e}} G.M. and Krohling, Renato A. and Ventura, Jos{\'{e}} A.}, year={2020}, month=Feb, pages={105162} }

@misc{typesafe2026models,
  author={{TypeSafe AI}},
  title={{Models: Jev 1.14 and Jev 1.13}},
  year={2026},
  howpublished={Model documentation},
  url={https://docs.typesafe.ai/models},
  note={Accessed 15 November 2026}
}

@misc{hendrycks2021imageneta,
      title={{Natural Adversarial Examples}}, 
      author={Dan Hendrycks and Kevin Zhao and Steven Basart and Jacob Steinhardt and Dawn Song},
      year={2021},
      eprint={1907.07174},
      archivePrefix={arXiv},
      primaryClass={cs.LG},
      url={https://arxiv.org/abs/1907.07174}, 
}

@misc{hendrycks2021imagenetr,
      title={{The Many Faces of Robustness: A Critical Analysis of Out-of-Distribution Generalization}}, 
      author={Dan Hendrycks and Steven Basart and Norman Mu and Saurav Kadavath and Frank Wang and Evan Dorundo and Rahul Desai and Tyler Zhu and Samyak Parajuli and Mike Guo and Dawn Song and Jacob Steinhardt and Justin Gilmer},
      year={2021},
      eprint={2006.16241},
      archivePrefix={arXiv},
      primaryClass={cs.CV},
      url={https://arxiv.org/abs/2006.16241}, 
}

@misc{hudson2019gqa,
      title={{GQA: A New Dataset for Real-World Visual Reasoning and Compositional Question Answering}}, 
      author={Drew A. Hudson and Christopher D. Manning},
      year={2019},
      eprint={1902.09506},
      archivePrefix={arXiv},
      primaryClass={cs.CL},
      url={https://arxiv.org/abs/1902.09506}, 
}

@misc{schwenk2022aokvqa,
      title={{A-OKVQA: A Benchmark for Visual Question Answering using World Knowledge}}, 
      author={Dustin Schwenk and Apoorv Khandelwal and Christopher Clark and Kenneth Marino and Roozbeh Mottaghi},
      year={2022},
      eprint={2206.01718},
      archivePrefix={arXiv},
      primaryClass={cs.CV},
      url={https://arxiv.org/abs/2206.01718}, 
}

@misc{gurari2018vizwiz,
      title={{VizWiz Grand Challenge: Answering Visual Questions from Blind People}}, 
      author={Danna Gurari and Qing Li and Abigale J. Stangl and Anhong Guo and Chi Lin and Kristen Grauman and Jiebo Luo and Jeffrey P. Bigham},
      year={2018},
      eprint={1802.08218},
      archivePrefix={arXiv},
      primaryClass={cs.CV},
      url={https://arxiv.org/abs/1802.08218}, 
}

@misc{liu2023mmbench,
      title={{MMBench: Is Your Multi-modal Model an All-around Player?}}, 
      author={Yuan Liu and Haodong Duan and Yuanhan Zhang and Bo Li and Songyang Zhang and Wangbo Zhao and Yike Yuan and Jiaqi Wang and Conghui He and Ziwei Liu and Kai Chen and Dahua Lin},
      year={2024},
      eprint={2307.06281},
      archivePrefix={arXiv},
      primaryClass={cs.CV},
      url={https://arxiv.org/abs/2307.06281}, 
}

@misc{fu2023mme,
      title={{MME: A Comprehensive Evaluation Benchmark for Multimodal Large Language Models}}, 
      author={Chaoyou Fu and Peixian Chen and Yunhang Shen and Yulei Qin and Mengdan Zhang and Xu Lin and Jinrui Yang and Xiawu Zheng and Ke Li and Xing Sun and Yunsheng Wu and Rongrong Ji and Caifeng Shan and Ran He},
      year={2025},
      eprint={2306.13394},
      archivePrefix={arXiv},
      primaryClass={cs.CV},
      url={https://arxiv.org/abs/2306.13394}, 
}

@misc{lu2024mathvista,
      title={{MathVista: Evaluating Mathematical Reasoning of Foundation Models in Visual Contexts}}, 
      author={Pan Lu and Hritik Bansal and Tony Xia and Jiacheng Liu and Chunyuan Li and Hannaneh Hajishirzi and Hao Cheng and Kai-Wei Chang and Michel Galley and Jianfeng Gao},
      year={2024},
      eprint={2310.02255},
      archivePrefix={arXiv},
      primaryClass={cs.CV},
      url={https://arxiv.org/abs/2310.02255}, 
}

@misc{lu2022scienceqa,
      title={{Learn to Explain: Multimodal Reasoning via Thought Chains for Science Question Answering}}, 
      author={Pan Lu and Swaroop Mishra and Tony Xia and Liang Qiu and Kai-Wei Chang and Song-Chun Zhu and Oyvind Tafjord and Peter Clark and Ashwin Kalyan},
      year={2022},
      eprint={2209.09513},
      archivePrefix={arXiv},
      primaryClass={cs.CL},
      url={https://arxiv.org/abs/2209.09513}, 
}

@misc{li2023seedbench,
      title={{SEED-Bench: Benchmarking Multimodal LLMs with Generative Comprehension}}, 
      author={Bohao Li and Rui Wang and Guangzhi Wang and Yuying Ge and Yixiao Ge and Ying Shan},
      year={2023},
      eprint={2307.16125},
      archivePrefix={arXiv},
      primaryClass={cs.CL},
      url={https://arxiv.org/abs/2307.16125}, 
}

@misc{masry2022chartqa,
      title={{ChartQA: A Benchmark for Question Answering about Charts with Visual and Logical Reasoning}}, 
      author={Ahmed Masry and Do Xuan Long and Jia Qing Tan and Shafiq Joty and Enamul Hoque},
      year={2022},
      eprint={2203.10244},
      archivePrefix={arXiv},
      primaryClass={cs.CL},
      url={https://arxiv.org/abs/2203.10244}, 
}

@misc{methani2020plotqa,
      title={{PlotQA: Reasoning over Scientific Plots}}, 
      author={Nitesh Methani and Pritha Ganguly and Mitesh M. Khapra and Pratyush Kumar},
      year={2020},
      eprint={1909.00997},
      archivePrefix={arXiv},
      primaryClass={cs.CV},
      url={https://arxiv.org/abs/1909.00997}, 
}

@misc{kafle2018dvqa,
      title={{DVQA: Understanding Data Visualizations via Question Answering}}, 
      author={Kushal Kafle and Brian Price and Scott Cohen and Christopher Kanan},
      year={2018},
      eprint={1801.08163},
      archivePrefix={arXiv},
      primaryClass={cs.CV},
      url={https://arxiv.org/abs/1801.08163}, 
}

@misc{kembhavi2016ai2d,
      title={{A Diagram Is Worth A Dozen Images}}, 
      author={Aniruddha Kembhavi and Mike Salvato and Eric Kolve and Minjoon Seo and Hannaneh Hajishirzi and Ali Farhadi},
      year={2016},
      eprint={1603.07396},
      archivePrefix={arXiv},
      primaryClass={cs.CV},
      url={https://arxiv.org/abs/1603.07396}, 
}

@misc{mathew2021docvqa,
      title={{DocVQA: A Dataset for VQA on Document Images}}, 
      author={Minesh Mathew and Dimosthenis Karatzas and C. V. Jawahar},
      year={2021},
      eprint={2007.00398},
      archivePrefix={arXiv},
      primaryClass={cs.CV},
      url={https://arxiv.org/abs/2007.00398}, 
}

@misc{singh2019textvqa,
      title={{Towards VQA Models That Can Read}}, 
      author={Amanpreet Singh and Vivek Natarajan and Meet Shah and Yu Jiang and Xinlei Chen and Dhruv Batra and Devi Parikh and Marcus Rohrbach},
      year={2019},
      eprint={1904.08920},
      archivePrefix={arXiv},
      primaryClass={cs.CL},
      url={https://arxiv.org/abs/1904.08920}, 
}

@inproceedings{mishra2019ocrvqa, title={{OCR-VQA: Visual Question Answering by Reading Text in Images}}, url={http://dx.doi.org/10.1109/ICDAR.2019.00156}, DOI={10.1109/icdar.2019.00156}, booktitle={2019 International Conference on Document Analysis and Recognition (ICDAR)}, publisher={IEEE}, author={Mishra, Anand and Shekhar, Shashank and Singh, Ajeet Kumar and Chakraborty, Anirban}, year={2019}, month=sep, pages={947--952} }

@inproceedings{mathew2022infographicvqa, title={{InfographicVQA}}, url={http://dx.doi.org/10.1109/WACV51458.2022.00264}, DOI={10.1109/wacv51458.2022.00264}, booktitle={2022 IEEE/CVF Winter Conference on Applications of Computer Vision (WACV)}, publisher={IEEE}, author={Mathew, Minesh and Bagal, Viraj and Tito, Ruben and Karatzas, Dimosthenis and Valveny, Ernest and Jawahar, C. V.}, year={2022}, month=Jan, pages={1697--1706} }

@InProceedings{johnson2017clevr,

author = {Johnson, Justin and Hariharan, Bharath and van der Maaten, Laurens and Fei-Fei, Li and Lawrence Zitnick, C. and Girshick, Ross},

title = {{CLEVR: A Diagnostic Dataset for Compositional Language and Elementary Visual Reasoning}},

booktitle = {Proceedings of the IEEE Conference on Computer Vision and Pattern Recognition (CVPR)},

month = {July},

year = {2017}

}

@misc{acharya2019tallyqa,
      title={{TallyQA: Answering Complex Counting Questions}}, 
      author={Manoj Acharya and Kushal Kafle and Christopher Kanan},
      year={2018},
      eprint={1810.12440},
      archivePrefix={arXiv},
      primaryClass={cs.CV},
      url={https://arxiv.org/abs/1810.12440}, 
}

@inproceedings{suhr2019nlvr2,
    title = "A Corpus for Reasoning about Natural Language Grounded in Photographs",
    author = "Suhr, Alane  and
      Zhou, Stephanie  and
      Zhang, Ally  and
      Zhang, Iris  and
      Bai, Huajun  and
      Artzi, Yoav",
    editor = "Korhonen, Anna  and
      Traum, David  and
      M{\`a}rquez, Llu{\'i}s",
    booktitle = "Proceedings of the 57th Annual Meeting of the {A}ssociation for {C}omputational {L}inguistics",
    month = jul,
    year = "2019",
    address = "Florence, Italy",
    publisher = "Association for Computational Linguistics",
    url = "https://aclanthology.org/P19-1644/",
    doi = "10.18653/v1/P19-1644",
    pages = "6418--6428"
}

@article{liu2023vsr,
    title = "Visual Spatial Reasoning",
    author = "Liu, Fangyu  and
      Emerson, Guy  and
      Collier, Nigel",
    journal = "Transactions of the Association for Computational Linguistics",
    volume = "11",
    year = "2023",
    address = "Cambridge, MA",
    publisher = "MIT Press",
    url = "https://aclanthology.org/2023.tacl-1.37/",
    doi = "10.1162/tacl_a_00566",
    pages = "635--651"
}

@inproceedings{bossard2014food101, title={{Food-101 -- Mining Discriminative Components with Random Forests}}, ISBN={9783319105994}, ISSN={1611-3349}, url={http://dx.doi.org/10.1007/978-3-319-10599-4_29}, DOI={10.1007/978-3-319-10599-4_29}, booktitle={Computer Vision -- ECCV 2014}, publisher={Springer International Publishing}, author={Bossard, Lukas and Guillaumin, Matthieu and Van Gool, Luc}, year={2014}, pages={446--461} }

@inproceedings{nilsback2008flowers102, title={{Automated Flower Classification over a Large Number of Classes}}, url={http://dx.doi.org/10.1109/ICVGIP.2008.47}, DOI={10.1109/icvgip.2008.47}, booktitle={2008 Sixth Indian Conference on Computer Vision, Graphics \& Image Processing}, publisher={IEEE}, author={Nilsback, Maria-Elena and Zisserman, Andrew}, year={2008}, month=Dec, pages={722--729} }

@misc{xai2024realworldqa,
  author={{xAI}},
  title={{RealWorldQA}},
  year={2024},
  howpublished={Dataset},
  url={https://huggingface.co/datasets/xai-org/RealworldQA},
  note={Released with Grok-1.5 Vision Preview, 12 April 2024; accessed 15 November 2026}
}

@misc{qwen2026qwen3827b,
  title={{Qwen3.8-27B} Model Card},
  author={{Qwen Team}},
  year={2026},
  howpublished={Model documentation},
  url={https://huggingface.co/Qwen/Qwen3.8-27B},
  note={Exact evaluated checkpoint family; accessed 15 November 2026}
}

@misc{gemmateam2026gemma4,
  title={{Gemma 4 Technical Report}},
  author={{Gemma Team}},
  year={2026},
  eprint={2607.02770},
  archivePrefix={arXiv},
  primaryClass={cs.CL},
  url={https://arxiv.org/abs/2607.02770}
}

@misc{google2026gemma431bit,
  title={{Gemma 4 31B IT} Model Card},
  author={{Google DeepMind}},
  year={2026},
  howpublished={Model documentation},
  url={https://huggingface.co/google/gemma-4-31B-it},
  note={Instruction-tuned checkpoint; accessed 15 November 2026}
}

@misc{d3_2026,
  title={{d3}: A Multimodal Foundation Decision Model},
  author={{vLLM Semantic Router Team}},
  year={2026},
  howpublished={\url{https://huggingface.co/vllm-sr/d3}}
}

@misc{autotrust2026deciderprovision27b,
  title={{AutoTrust-Decider-Pro-Vision-27B} Model Card},
  author={{AutoTrust AI Lab}},
  year={2026},
  howpublished={Model documentation},
  url={https://huggingface.co/autotrust/AutoTrust-Decider-Pro-Vision-27B},
  note={Current repository reached from autotrust/JEV-27B-VL; accessed 15 November 2026}
}

@misc{li2026soarena,
  title={{SO-Arena}: Benchmarking the Capability Spectrum and Reliability of System One Models},
  author={Li, Lichen and Dong, Ziping and Feng, Sicheng and Li, Qi and Wang, Xinchao},
  year={2026},
  eprint={2611.15777},
  archivePrefix={arXiv},
  primaryClass={cs.CL},
  url={https://arxiv.org/abs/2611.15777}
}
